ORIGINAL ARTICLE

# Digitally enriching a high-risk population for pancreatic cancer using routine blood-based measures and clinical histories

**Authors and Affiliations**

Chris Varghese[1,2*], Leo Y. Li-Han[1*], Richa Bisht[1], Ellen Larson[1], Frank Lee[1], Ryan M. Carr[1], Tanios S. Bekaii-Saab[3], Shounak Majumder[4], John D. Halamka[5], Mark Truty[1], Ajit H. Goenka[6], Hojjat Salehinejad[7,8], Cornelius A. Thiels[1]

*Co-First Authorship

1. Department of Surgery, Mayo Clinic, Rochester, MN, USA
2. Department of Surgery, University of Auckland, Auckland, NZ
3. Department of Hematology and Oncology, Mayo Clinic, Phoenix, AZ, USA
4. Division of Gastroenterology and Hepatology, Mayo Clinic, Rochester, MN, USA
5. Mayo Clinic Platform, Mayo Clinic, Rochester, Minnesota
6. Department of Radiology, Mayo Clinic, Rochester, Minnesota
7. Division of Health Care Delivery Research, Robert D. and Patricia E. Kern Center for the Science of Health Care Delivery, Mayo Clinic, Rochester, MN, USA
8. Department of Artificial Intelligence and Informatics, Mayo Clinic, Rochester, MN, USA

**Corresponding Author**

Cornelius A Thiels DO, MBA
Division of Hepatobiliary and Pancreas Surgery
Department of Surgery
Mayo Clinic Rochester
200 First St SW, Rochester, MN 55902
Phone: (507)284-2095
E-mail: thiels.cornelius@mayo.edu

**Contributors**

CV, LL, and CT designed the study. HS and CT provided supervision. CV and LL contributed to code writing, model training, and data analysis. All authors wrote the first draft. All authors revised the manuscript. All authors had access to the result data presented in the final manuscript and all authors read and approved the final manuscript for submission.

**Conflicts of Interest/Disclosures**

No authors declare related financial disclosures.

- CV is a member of Alimetry Ltd and has IP in gastrointestinal motility diagnostics.

- CT is an inventor of Mayo Clinic intellectual property which is licensed to apoQlar MEdical andf HoloMedX LLC and may receive royalties paid to Mayo Clinic and serves on advisory board for Boston Scientific.
- SM is an inventor of Mayo Clinic IP which is licensed to Abbott Cancer Diagnostics (formerly Exact Sciences) and may receive royalties paid to Mayo Clinic.

The remaining authors declare no commercial or financial conflicts of interest.

**Funding/Support**
This work was funded by the Mayo Clinic Presidents Strategic Initiative Fund. The following authors declare additional funding received, unrelated to this project:

- CT receives grant funding from the Mayo Clinic Presidents Strategic Initiative Fund, Mayo Clinic Center for Digital Health Dalio Foundation AI/ML Enablement Award, Mayo Clinic Simons Family Career Development Award in Surgical Innovation, and Johnson and Johnson Polyphonic AI Fund for Surgery QuickFire Challenge.
- CV receives funding from the Health Research Council of New Zealand Clinical Research Training Fellowship and Artificial Intelligence (AI) in Healthcare Project Grant.
- AHG receives grant funding from the National Institutes of Health (Grant Numbers R01CA272628 and R01CA256969), Mayo Clinic Comprehensive Cancer Center, Champions for Hope Pancreatic Cancer Research Program of the Funk Zitiello Foundation, David E. Simon & Jacqueline S. Simon Charitable Foundation, and Hoveida Family Foundation.

**Data sharing**
The datasets are derived from de-identified cohorts within the Mayo Clinic Platform (https://www.mayoclinicplatform.org/). Code generated during the project is available at: https://github.com/lcapacitor/premod. The trained model is not publicly available, but can be made available for external validation, with appropriate data use and privacy agreements. This can be requested from the corresponding author.

**Acknowledgments**
We acknowledge the Mayo Clinic Platform team for their technical support.

**Word count:** 3312

## Abstract

### Importance

Earlier detection of pancreatic cancer is key to enabling wider access to curative treatment and reducing cancer deaths; however, screening is presently not viable.

### Objective

We aimed to use latent digital indicators to predict the risk of pancreatic cancer.

### Design

Longitudinal sequences of coded diagnoses and blood test values accrued by patients throughout their clinical interactions were used to train a custom Transformer-based neural network with a multi-head attention mechanism to predict risk of pancreatic cancer with a multi-year lead time and risk-stratify populations for targeted screening.

### Setting

Mayo Clinic Platform with trained model from Mayo Clinic Rochester validated at Mayo Clinic Arizona, Mayo Clinic Florida, and the centers aggregated within the Mayo Clinic Health Systems.

### Participants

The cohort comprised 6,017 adults with pancreatic cancer and 177,081 controls (overall median age 75, 45% female) with median 12 years (interquartile range 6.9-16.2) of medical history prior to pancreatic cancer diagnosis.

### Exposure

Pancreatic cancer diagnosis at a lead time of 1-, 2-, and 3-years prior to diagnosis.

### Main Outcomes and Measures

Area under the receiver operating characteristic (AUROC), calibration curve slope and intercept, Brier score, and diagnostic odds ratio.

**Results**

External validation via leave-one-site-out, out-of-sample testing predicting pancreatic cancer 1-, 2-, and 3-years prior to diagnosis demonstrated mean area under the receiver operating characteristic of 0.837 (95% confidence interval 0.827-0.848), 0.797 (95% confidence interval 0.782-0.813), and 0.760 (95% confidence interval 0.745-0.776), respectively. Estimated pancreatic cancer risks were well-calibrated (calibration plot slope 1.08, intercept of -0.077; Brier score 0.025), and a Bayesian update based on the population prevalence of pancreatic cancer allows estimated cancer risk outputs to be transportable across settings. At testing, a screening threshold of >3.3% risk of pancreatic cancer in 1-year offered a diagnostic odds ratio of 18.2.

**Conclusions and Relevance**

We present a digital population-level risk enrichment tool that could justify scaling of pancreatic cancer screening and widen access to curative-intent management.

## Introduction

Pancreatic cancer is projected to become the second leading cause of cancer-related death in the USA by 2030.[1] A diagnosis of pancreatic cancer is lethal for most patients, with median survival of 4 months, and 5-year survival of 13%.[2] Poor outcomes are largely driven by the late stage of clinical presentation, consequent to a clinically-occult phase of incident disease progression.[3] Enabling early detection of pancreatic cancer at the population level could therefore broaden access to curative treatment options and improve outcomes for this otherwise recalcitrant cancer.

Unlike malignancies where screening initiatives target well-defined premalignant lesions or established oncogenic drivers; pancreatic cancer lacks actionable biomarkers that reliably signal early-stage disease.[4] Even premalignant pancreatic cysts are only seen in <10% of pancreatic cancers.[5] Currently, population-level screening is economically prohibitive as pancreatic cancer is rare, can evolve rapidly from visually occult to unresectable disease in short time frame,[6,7] and current screening modalities such as endoscopic ultrasound and magnetic resonance imaging are highly expensive, resource-intensive, and impractical for longitudinal surveillance at scale.[8] Whilst radiologic surveillance is performed among individuals with recognized risk factors for pancreas cancer, including genetic susceptibility,[9] family history, and potentially new-onset diabetes,[10] this misses the majority of pancreatic cancer which lack these established risk factors. Hence, inability to risk stratify the population for incident pancreatic cancer constrains the implementation of targeted screening for early disease.[11]We hypothesize that these pragmatically acquired, informative, and biologically plausible indicators of malignancy risk can be operationalized to digitally risk-stratify individuals for pancreatic cancer.

Here, leveraging a multi-institutional platform,[23] we present our multimodal Transformer that integrates routinely measured blood-based biomarkers and individual clinical histories to identify early, pre-diagnostic signatures of pancreatic cancer up to 1-, 2-, and 3-years prior to diagnosis.

## Methods

*<u>Data source, reporting, and sharing</u>*

This study was approved by the Mayo Clinic Institutional Review Board (25-010099) and involves analysis of de-identified data via the Mayo Clinic Platform_Discover (Platform).[23] The Platform is HIPAA compliant. Any data beyond what is reported in the manuscript, including but not limited to the raw EHR data, cannot be shared or released due to the parameters of the expert determination to maintain the data de-identification. This manuscript is reported in accordance with the STrengthening the Reporting of OBservational studies in Epidemiology (STROBE) and Transparent Reporting of a multivariable prediction model for Individual Prognosis Or Diagnosis; AI (TRIPOD+AI) guidelines.[24,25]

*<u>Population</u>*

All longitudinal diagnostic and blood-based biomarker trajectories used in this study were obtained from the Mayo Clinic Platform Discovery, which contains longitudinal data from the three tertiary and quaternary medical centers and the Mayo Clinic Health Systems (MCHS) which combines data from 17 different community hospitals. Patients with at least one diagnosis code in ICD chapters of 157 (ICD-9) and C25 (ICD-10) were identified as pancreatic cancer cases. A sensitivity analysis of pancreatic cancer diagnosis was performed in a cohort of patients confirmed to have pancreatic cancer based on a large-language model-based synthesis of radiology, pathology, and clinical notes (**Supplementary Methods; Sensitivity analysis of with further diagnostic confirmation**). The prediction target was risk of future recorded pancreatic cancer diagnosis. The date of the first cancer diagnosis was defined as the patient's diagnosis date. All diagnosis codes and blood-based biomarkers within the 20 years preceding the patient's diagnosis date were included in the analysis. A list of the predefined 31 blood tests as well as their histograms are provided in **Supplementary Figure 1**. These blood tests were selected based on literature-founded or clinically putative associations with malignancy. Patients with no results for any of the 31 blood tests of interest were excluded. Other exclusion criteria adopted in the study include (1) patients less than 18 years at the diagnosis date, (2) patients with less than 3 years of pre-cancer diagnostic history, (3) patients with fewer than 3 measurements for any blood tests of interest.

For controls, patients with none of the following 20 diagnosis codes, including pancreatic, esophagus, stomach, gallbladder, liver, small and large intestines, were identified: C15, C16, C17, C18, C19, C20, C21, C25, D00, D01, C7A, 150, 151, 152, 153, 154, 155, 156, 157, and 230. We excluded these other cancers as separate work is currently ongoing to develop a multicancer prediction tool. We defined the date one year before their most recent record as the “diagnosis date” (to avoid uncertainty of undiagnosed cancer prior to the end of a record or death), and up to 20 years of pre-diagnosis data were retrospectively extracted following the same criteria, per the methods of Placido et al.[15] These controls were age- and sex-matched to cases with pancreatic cancer. **Supplementary Figure 2** illustrates the cohort selection flowcharts for case and control groups. Therefore, a total of 183,098 patients (case:control = 6017:177,081; 1:29) were included in the study.

*Analysis*

Irregularly sampled time series data including ICD diagnosis codes and blood test values were represented using time-bucketed frequency encoding (**Supplementary Methods; Time series data representation**). Our unique time-bucketed data organization structure inherently handles irregular and heterogeneous medical history and blood test acquisition histories, such that missing data do not hinder training, validation, or deployment utility of our model. This approach simplifies the integration of many blood-based markers while minimizing short-scale noise (e.g., from acute illnesses or measurement error). **Figure 1** summarizes our Transformer-based architecture incorporating this time bucketed preprocessing and attention-based feature aggregation.

**Figure 1B** demonstrates the overall architecture of the proposed pancreatic cancer risk detection model (‘PRE-diagnostic Pancreatic Cancer Risk MODel’; PREMOD). Following the sequential modeling paradigm, the model was developed based on the Transformer encoder architecture,[27] which comprises four components: input encoder, feature extractor, feature aggregator, and classifier. Modeling is further detailed in the **Supplementary Methods; Model**.

To maximize data usage and have performance metrics reflect real-world generalizable performance, we used a leave-one-site-out strategy across three different sites, namely Mayo Clinic Health Systems (MCHS; 17 community hospitals), a tertiary multicampus Arizona medical center, and a tertiary multicampus Florida hospital, with each site iteratively used as the independent testing set. Each patient was only included once in each site based on their location of diagnosis. Further 10-fold cross-validation was applied to the remaining data (development set), for model training, validation, and hyperparameter tuning (further detailed in **Supplementary Methods; Training and evaluation**). These methods comply with the latest predictive model development guidelines promoting maximal utilization of available data whilst simultaneously informing on external generalizability.[27]

Model discrimination was assessed via area under the receiver operating characteristic (AUROC); calibration via slope and intercept of the calibration plot, and the expected calibration error (ECE); and clinical utility through the diagnostic odds ratio (DOR), and pancreatic cancer screening simulations (further detailed in the **Supplementary Methods; Training and evaluation and Modelling population-level screening feasibility**). Output probabilities were recalibrated using a Bayesian prevalence-aware recalibration approach that enables generalization of model performance across various prevalence settings (see **Supplementary Methods; Bayesian prevalence-aware recalibration**). Continuous risk assessment was modelled using a temporal sliding window approach (**Supplementary Methods; Simulation of real-world continuous risk prediction**). Sensitivity, specificity, positive predictive value (PPV), and negative predictive value (NPV) were interrogated across clinical decision thresholds, including via data-driven methods such as maximization of the Younden’s J index (a metric that considers both sensitivity and specificity). Decision curve analysis was used to analyze clinical utility.

## Results

*Cohort*

Our multi-institutional cohort included centers caring for patients with pancreatic cancer across a range of settings from community-focused clinics to quaternary centers delivering advanced care (summarized in **Supplemental Table 1**). Our cohort comprised 183,098 patients, (median age 75) from 4 geographic centers (Rochester: n=84,350, Mayo Clinic Health Systems [MCHS]: n=52,218, Arizona: n=23,588, and Florida: n=22,942), of whom 6017 (3%) were diagnosed with pancreatic cancer. Overall, patients had median 12.0 years of clinical histories available, ranging from January 1985 to April 2024, with median 192 diagnostic codes and median 158 blood test values per patient. For individual centers, clinical history data remained similar with 12.8 years of data (median 145 diagnosis codes and 123 blood test values) in Rochester, 13.5 years (307 diagnosis codes and 240 blood test values) in MCHS, 9.3 years (201 diagnosis codes and 149 blood test values) in Arizona, and 8.3 years (161 diagnosis codes and 162 blood test values) in Florida.

*Model performance with geographic cross-validation*

Using geographic leave-one-site-out cross validation, which leverages available data and simultaneously examines independent external performance in different healthcare settings, our Transformer model showed good to excellent generalizable performance.[29] Predicting pancreatic cancer risk with a 1-year lead time showed an AUROC of 0.872 (95% CI 0.866—0.879) in the community health systems, 0.830 (95% CI 0.824—0.835) in Arizona, 0.810 (95%CI 0.805—0.816) in Florida, and an overall average AUROC of 0.837 (95% CI 0.827—0.848). Performance remained robust when predicting with a 2-year lead time with an AUROC of 0.846 (95% CI 0.842—0.851) in the community health systems, 0.792 (95% CI 0.784—0.800) in Arizona, 0.754 (95% CI 0.740—0.768) in Florida, and an overall average AUROC of 0.797 (95% CI 0.782—0.813); and 3-year lead time with an AUROC of 0.804 (95% CI 0.795—0.814) in the community health systems, 0.768 (95% CI 0.761—0.774) in Arizona, 0.709 (95% CI 0.701—0.717) in Florida, and an overall average AUROC of 0.760 (95% CI 0.745—0.776). Receiver operating characteristic curves are shown in **Figure 2A-C**. Sensitivity analyses with LLM-confirmed pancreatic cancer diagnosis is reported in **Supplementary Table 2** and showed robust performance. Ablation analysis with diagnostic codes alone or blood tests values alone are reported in **Supplementary Figure 3**.

The estimated risk of pancreas cancer predicted by the model was well calibrated to observed pancreatic cancer rates as indicated by an average calibration plot slope of 1.075 (closer to 1 indicating optimal calibration), calibration plot intercept of -0.077 (closer to 0 indicating better calibration in the large), and an average expected calibration error (ECE) of 0.037 (closer to 0 indicating better calibration), with a 1-year prediction lead time. Slope was 1.034, intercept was -0.1, and ECE was 0.04 at 2-years prediction lead time, and 0.99, -0.12, and 0.051 at 3-years respectively. Evaluating site-wise calibration with the Brier score (measure of discrimination and calibration) confirms these results generalize (**Table 2**). This probabilistic accuracy is important to be able to interrogate risk thresholds at which population-level screening might become scalable and economically feasible (**Figure 2D-F**; see 'Screening' section below).

*Performance and risk classification trade-offs*

Using a MCHS training data prevalence of 3.3% as a high-risk threshold yields a sensitivity of 95.3%, specificity of 55.2%, PPV of 5.8%, and NPV of 99.8%. A more conservative threshold that maximizes the Youden's J index at a threshold of 6.6% offers a sensitivity of 85.9%, specificity of 69.6%, PPV of 7.5% and NPV of 99.4%, while an even more conservative threshold still at 50% risk, provides a sensitivity of 25.6%, specificity of 99.7%, PPV of 70.7% and NPV of 97.9%, with higher risk thresholds enabling better resource utilization for downstream screening and diagnostic investigations (**Supplemental Table 3**).

*Pancreatic cancer risk dynamically over time*

Our Transformer model principally derives its learning from the temporal patterns of healthcare interactions, specifically the acquisition of new and repeated diagnoses and blood test values. These are used to quantify an individual's pancreas cancer risk over time, showing risk trajectories across retrospective clinical histories (**Figure 3**). This temporal patterning indicates latent signals for pancreatic cancer become evident ~5 years ahead of diagnosis within the electronic medical record. Notably, in tertiary and quaternary pancreas centers including Rochester, Arizona, and Florida, the lead time with which over 20% pancreatic cancer risk could be identified is shorter than the ~5-year lead-time seen in the community setting (MCHS). This likely reflects the more advanced disease seen at these centers, with >20% risk estimated closer to 3-4 years prior to diagnosis in Arizona and Florida. Long range lead-times in the magnitude of >3 years offers advantages to detect the varying phenotypes of pancreatic cancer reported in the literature including those that evolve gradually followed by rapid metastases or the systematically punctuated equilibrium phenotype.[30]

*Biomedical plausibility evaluated through model explainability*

The top 20 most important features based on their marginal contributions to the model as evaluated by Shapley additive explanation values are shown in **Figure 4A-B**, largely consistent with known risk features from the electronic medical record for pancreatic cancer.[12,15,16,19] Of note, complete blood count parameters were particularly informative for pancreatic cancer risk based on explainability analysis, alongside diagnostic codes associated with metabolic disorders, including diabetes, health system interactions (e.g., for vaccinations), benign pancreatic conditions (e.g., pancreatitis), and liver function markers. The relative importance and contribution of these parameters over time as you approach within one year of pancreatic cancer diagnosis is shown in **Figure 4A-B**. Overall importance and temporal contributions of all international classification of disease chapter codes and blood test types are visualized in **Figure 4C-D**.

*Clinical utility analysis*

Decision Curve Analysis demonstrates that PREMOD provides greater net benefit than both a “Treat-All” (assuming all have pancreatic cancer) and “Treat-None” (assuming no one has pancreatic cancer) strategies across a wide spectrum of threshold probabilities, suggesting clear clinical utility. Because pancreatic cancer is a low-prevalence disease where invasive screening carries harms and cost, the "Treat All" strategy (green dotted line) rapidly loses value (**Supplementary Figure 4**). In contrast, at the 3.3% threshold used in the primary analysis, PREMOD achieved a net benefit of approximately 12 additional true-positive predictions per 1000 patients without generating new false-positives when predicting pancreatic cancer risk with a 1-year lead time. With the same threshold, the model's net benefit reached around 10 and 7 additional true-positive predictions per 1000 patients at the 2- and 3-year prediction lead times, respectively. Net benefit was seen across 10-85% risk threshold spectrum for predictions with a 1-year lead-time, 17-85% risk range for predictions with a 2-year lead-time, and 22-82% for 3-years.

**Discussion**

We present and validate a novel Transformer-based pancreatic cancer risk model across diverse healthcare settings, laying the foundation for a population-level pancreatic cancer risk enrichment program. Our discriminative and highly probabilistically accurate model offers a scalable population-based enrichment strategy to guide downstream screening and/or diagnostic investigations such as through CT, endoscopy or MRI, with potential to improve screening efficiency substantially. An era of early detection of pancreas cancer could now be made feasible through population-level risk enrichment enabled by latent digital signatures from medical histories and blood test trajectories. Such an advance introduces a new hope for improved diagnosis of pancreatic cancer with curative potential.

This work extends upon previous unimodal deep learning strategies by adding pathophysiological insight through blood-based biomarkers, and hence improving the biological plausibility of model learning.[41] Blood-based biomarkers also likely offer a richer additive contribution to our model's learning than potentially duplicative information provided by medication prescribing data, as supported by improved discriminative performance of our model.[16] We further extend previous work with a comprehensive evaluating of probabilistic accuracy through extensive calibration analysis. This is pivotal to ensure the veracity of threshold-dependent screening simulations. Use of contemporary geographic internal-external cross-validation analysis across diverse healthcare settings also extends our validation analyses beyond historic static train/validation/test split methods.[31] Further, our Bayesian update approach to calibration accuracy allows for model portability across different incidence settings without needing retraining like previous models, which is particularly important given the rapidly increasing rates of pancreatic cancer.[12,14,37]

A key feature of our present work is the use of standardized data inputs concordant with Observational Medical Outcomes Partnership (OMOP) Common Data Models, which supports model transportability across health systems.[32] This was enabled through the standardized, de-identified data available through the Mayo Clinic Platform.[23] Further, leveraging widely used frameworks such as International Classification of Diseases (ICD) codes, and OMOP-standardized coding of blood test results, we ensure that our model can be recreated in any OMOP-compliant data warehouse, allowing for robust reproducibility and lowered barrier to implementation across diverse settings. Our institution partners with the Coalition for Healthcare AI (CHAI) partner-organization, ensuring rigorous data provenance for AI model development, overcoming a key limitation of previous longitudinal disease prediction models.[33,34]

Our model overcomes a critical challenge in pancreas cancer screening, where general population enrichment for screening is difficult owing to the lack of reliable pancreatic cancer risk-stratification. Previous promising work, uses a phenomena unique to pancreatic cancer, that is new-onset diabetes,[12,14,35] to identify a suitably high-risk population for targeted screening; however this only occurs in 34-40% of new diagnoses, leading to non-applicability in the majority of sporadic pancreatic cancers.[12,36,37] With respect to premalignant lesions, <10% of pancreatic cancers have a precursor cyst lesion.[30] Furthermore, only 10% of pancreatic cancer is familial, of whom 20% have a hereditable and identifiable mutation.[10,38] To operationalize this genetic risk, the Cancer of Pancreas Screening (CAPS) consortium targeted individuals with >5% risk of pancreatic cancer for annual EUS or MRI screening, identifying 10 cancers in among 1,461 high-risk individuals.[20,22,47] Hence, current efforts to enrich the general population for screening misses a substantial proportion of pancreatic cancers, necessitating a more sensitive approach.

We propose this model could be implemented to support a more economically feasible screening framework with broader applicability to a wider range of patients at risk of sporadic pancreatic cancer, including the >60% of pancreas cancer patients without a diabetes signature, and >90% of pancreas cancer patients without a familial risk factor.[10,13,14,38] One such proposal includes combining our model tuned to a high-sensitivity threshold ahead of a higher specificity, downstream diagnostic test, such as a CT scan, including with emerging CT-based radiomic models able to detect visually occult pancreas lesions such as REDMOD.[30] Use of our PREMOD digital screen significantly improves the number needed to screen to identify a pancreatic cancer. Importantly, this offers screening to a substantially increased proportion of the population compared to current standards, offering newfound hope for a lethal and rapidly progressing cancer, where early detection of surgically curative disease remains the only strategy for progress.

The opportunity for early detection through multi-year lead times compared to current diagnostic paradigms, paired with the potential to detect visually occult lesions may enable the breakthrough concept of pancreatic cancer interception.[20] KRAS activating mutations are highly prevalent in pancreatic ductal adenocarcinoma, and promising early results of KRAS inhibition may open the door to preventative therapy.[20,22,39] A critical challenge to realizing these benefits is identifying pancreatic cancer at the clinically occult, asymptomatic, precursor lesion-stage. Our digital screening approach offers a high-accuracy modality that can identify elevated risk up with up to 5-years lead time, offering a unique opportunity to target early-stage disease.

This study is not without limitations. Despite robust validation experiments across diverse healthcare settings, all data arise from Mayo Clinic hospitals and hence further testing is warranted across diverse patient groups and health systems. We also acknowledge that this study does not use a population-level cohort, however this is mitigated with the use of our Bayesian risk update methodology that ensures predictions from our case-control study are transportable to real-world low prevalence setting, indicating methodological readiness for prospective testing compared to previous models.[15,16] An important limitation of the current model's premise is the requirement for healthcare interactions to collect datapoints as inputs to the model, showing value of the 'healthcare interactome'. This may, however, disadvantage individuals of low socioeconomic background and limited access to healthcare.[40] Furthermore, at present, reliable staging data at diagnosis was not extractable from the Mayo Clinic Platform, and stage-shift is assumed based on robust predictive lead-times. This data is being pursued through large-language model augmented structured data abstraction for further model validation and testing.

Strengths of this work include the internal-external geographic cross-validation, and novel calibration optimization strategies that augment transportability of pancreatic cancer risk estimates. The next steps toward advancing our digital enrichment tool to guide pancreatic cancer screening includes prospective validation studies with assessment of stage-shift, lead-time-adjusted survival, and screening cost-effectiveness analysis based on life-years gained.

In conclusions, we present a discriminative and well calibrated model for pancreatic cancer risk with up to a 3-year lead time. Our model reliably identifies patients at high-risk of pancreatic cancer, which through digital population enrichment, lays the foundation for population level screening targeting early detection and increased curative-intent management of pancreatic cancer.

## Tables

**Table 1:** Discrimination and calibration performance of the model across sites and prediction lead-times.

| Prediction Lead Time | Center | AUROC | Brier Score | Calibration Slope | Calibration Intercept | Expected Calibration Error |
|---|---|---|---|---|---|---|
| **1-year** | MCHS | 0.872 [0.866, 0.879] | 0.023 | +1.061 | -0.098 | 0.033 |
| | Arizona | 0.830 [0.824, 0.835] | 0.026 | +1.145 | -0.059 | 0.035 |
| | Florida | 0.810 [0.805, 0.816] | 0.025 | +0.941 | -0.038 | 0.043 |
| | Overall | 0.837 [0.827, 0.848] | 0.025 | +1.075 | -0.077 | 0.037 |
| **2-year** | MCHS | 0.846 [0.842, 0.851] | 0.025 | +1.037 | -0.113 | 0.041 |
| | Arizona | 0.792 [0.784, 0.800] | 0.028 | +1.142 | -0.085 | 0.041 |
| | Florida | 0.754 [0.740, 0.768] | 0.028 | +0.847 | -0.059 | 0.048 |
| | Overall | 0.797 [0.782, 0.813] | 0.027 | +1.034 | -0.100 | 0.043 |
| **3-year** | MCHS | 0.804 [0.795, 0.814] | 0.028 | +1.001 | -0.136 | 0.050 |
| | Arizona | 0.768 [0.761, 0.774] | 0.030 | +1.122 | -0.113 | 0.048 |
| | Florida | 0.709 [0.701, 0.717] | 0.030 | +0.826 | -0.073 | 0.056 |
| | Overall | 0.760 [0.745, 0.776] | 0.029 | +0.991 | -0.119 | 0.051 |

AUROC, area under the receiver operating characteristic curve; MCHS, Mayo Clinic Health Systems. AUROC values are presented with 95% confidence intervals.

## Figures

**Figure 1:** (A) Visualization of time-bucketed data encoding matrices of longitudinal ICD diagnosis codes and laboratory tests for two subjects from case and control groups. (B) Architecture of the Transformer-based pancreatic cancer prediction model. (C) The model's output includes a pancreatic cancer score and a binary prediction result. (D) Simulation of pancreatic cancer risk continuous monitoring in real-world applications using temporally windowed data.

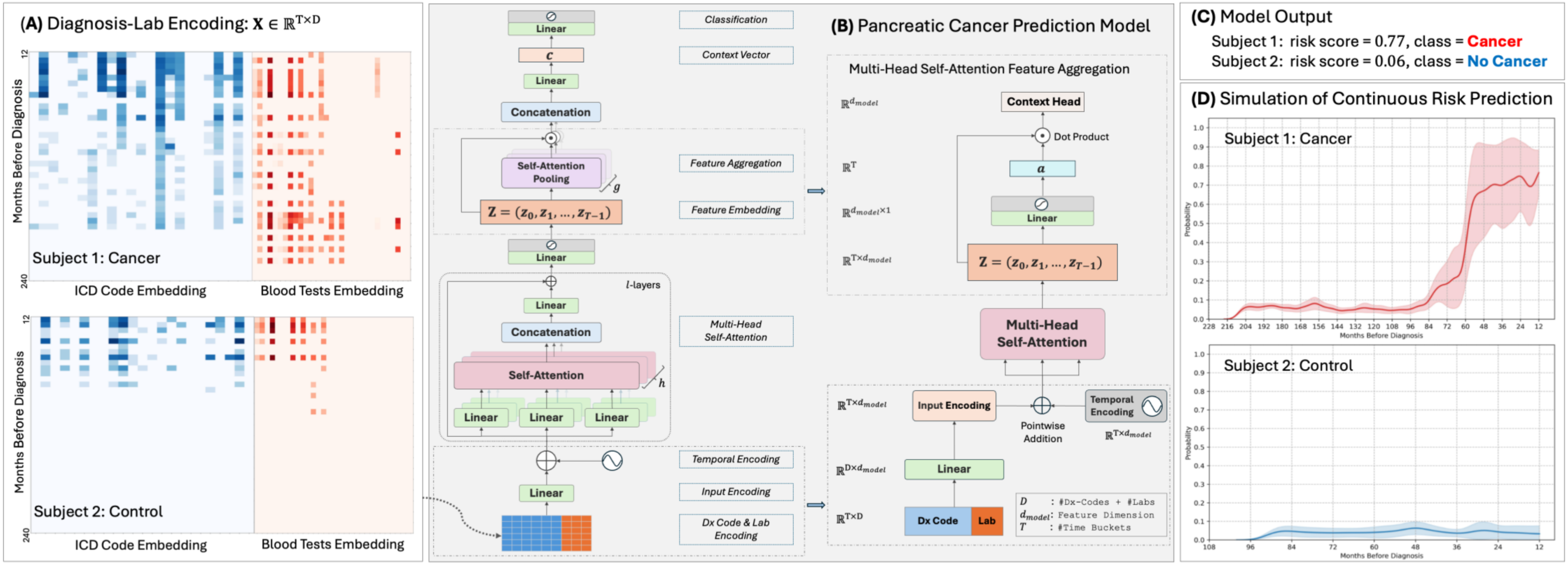

**Figure 2**: (A)—(C) Receiver operating characteristic curves for the models with the 1-, 2-, and 3-year prediction lead time across geographic centers (data are presented as mean±standard deviation). (D)—(F) Calibration curves for the models with the 1-, 2-, and 3-year prediction lead time, plotted with 95% confidence intervals.

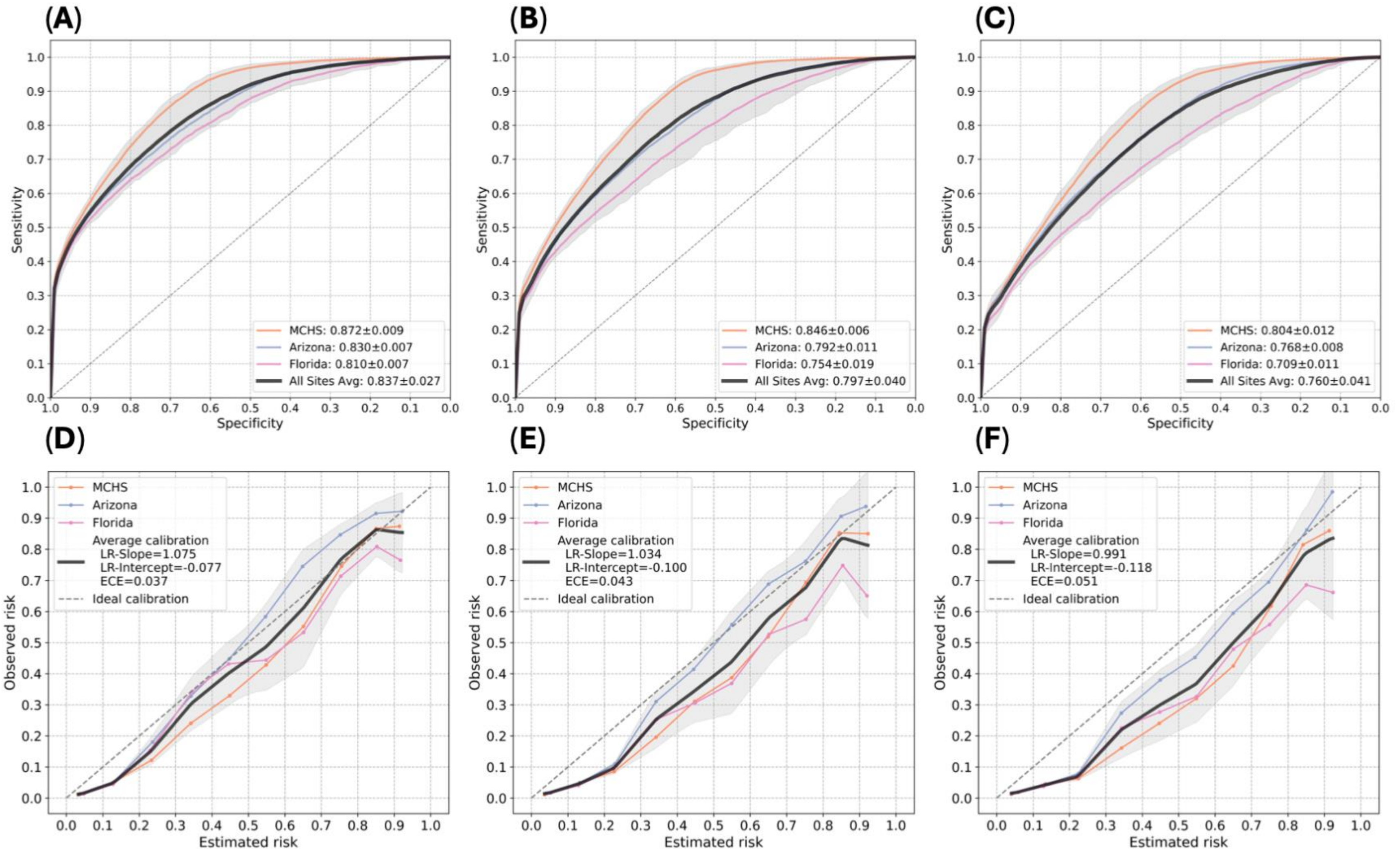

**Figure 3**: Model's predicted pancreatic cancer risk for case (red) and control (blue) groups over time in the geographic centers of (A) Rochester, (B) Mayo Clinic Health System, (C) Arizona, and (D) Florida.

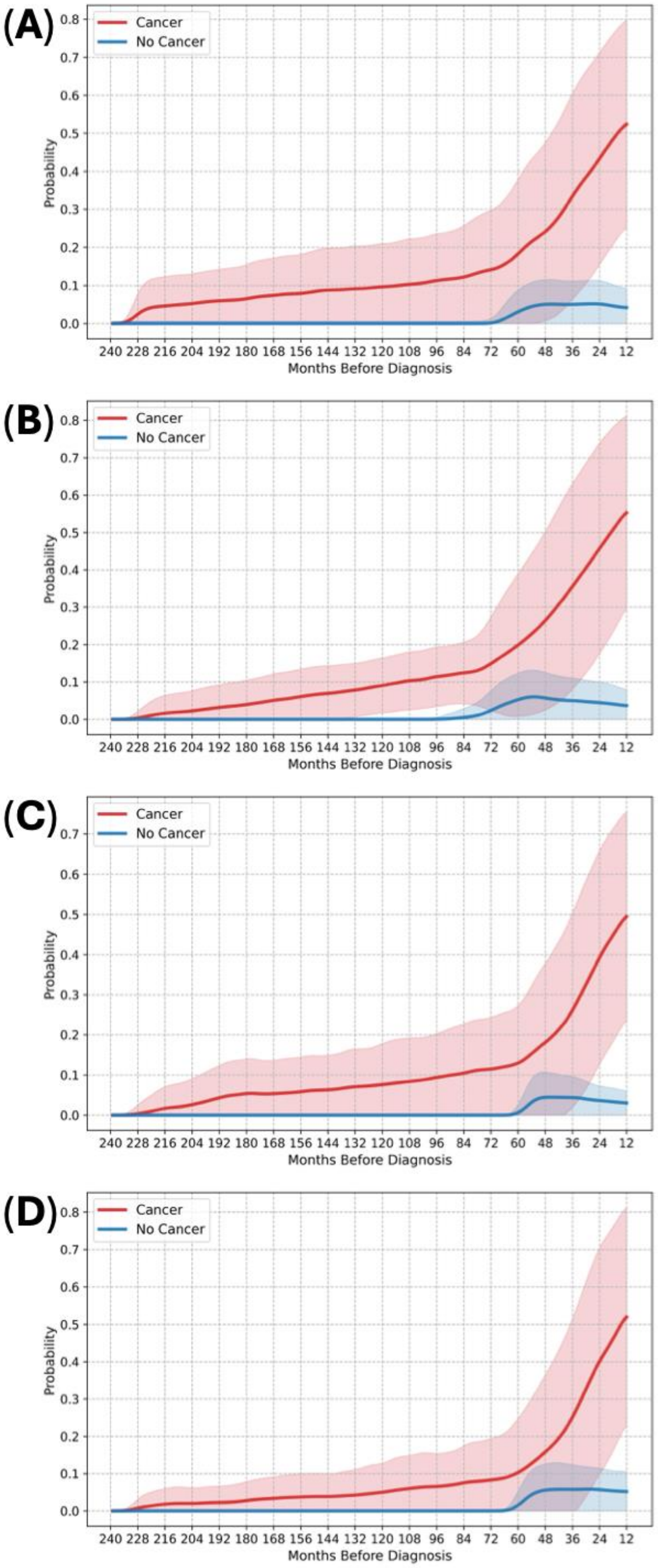

**Figure 4:** Top 20 features (ICD codes and blood tests) contributing to pancreatic cancer detection with their (A) contributions over time with bubble size representing contribution magnitudes and the (B) overall contributions aggregated by time based on SHAP values. Pancreatic cancer detection contributions of all features grouped by ICD diagnosis code chapters and laboratory test groups are shown over time in (C) and aggregated over time in (D).

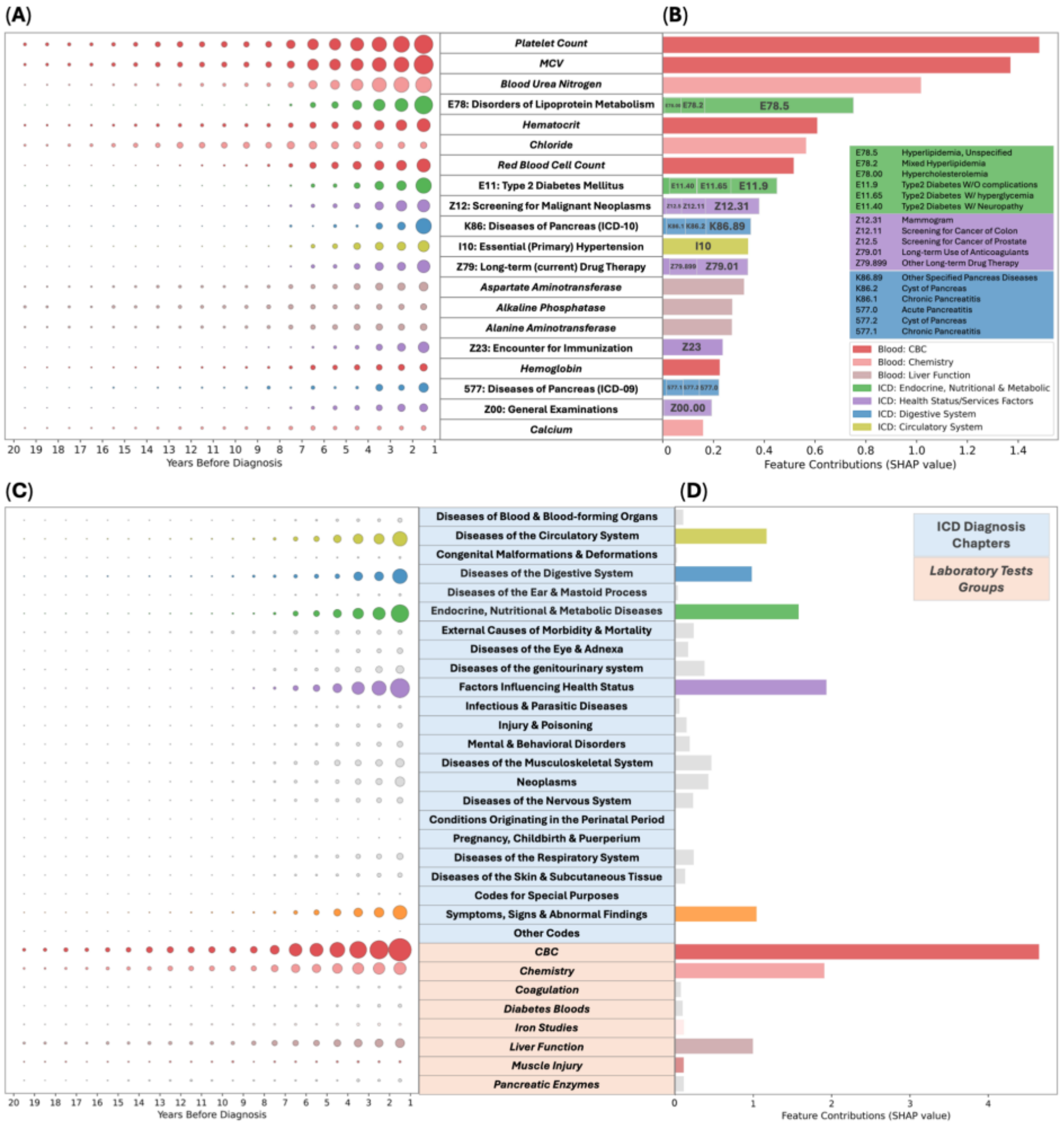

SUPPLEMENTARY APPENDIX

# Digitally enriching a high-risk population for pancreatic cancer using routine blood-based measures and clinical histories

**Authors and Affiliations**
Chris Varghese[1,2*], Leo Y. Li-Han[1*], Richa Bisht[1], Ellen Larson[1], Frank Lee[1], Ryan M. Carr[1], Tanios S. Bekaii-Saab[3], Shounak Majumder[4], John D. Halamka[5], Mark Truty[1], Ajit H. Goenka[6], Hojjat Salehinejad[7,8], Cornelius A. Thiels[1]

*Co-First Authorship

1. Department of Surgery, Mayo Clinic, Rochester, MN, USA
2. Department of Surgery, University of Auckland, Auckland, NZ
3. Department of Hematology and Oncology, Mayo Clinic, Phoenix, AZ, USA
4. Division of Gastroenterology and Hepatology, Mayo Clinic, Rochester, MN, USA
5. Mayo Clinic Platform, Mayo Clinic, Rochester, Minnesota
6. Department of Radiology, Mayo Clinic, Rochester, Minnesota
7. Division of Health Care Delivery Research, Robert D. and Patricia E. Kern Center for the Science of Health Care Delivery, Mayo Clinic, Rochester, MN, USA
8. Department of Artificial Intelligence and Informatics, Mayo Clinic, Rochester, MN, USA

**Corresponding Author:**
Cornelius A Thiels DO, MBA
Division of Hepatobiliary and Pancreas Surgery
Department of Surgery
Mayo Clinic Rochester
200 First St SW, Rochester, MN 55902
Phone: (507)284-2095
E-mail: thiels.cornelius@mayo.edu

## Supplementary Methods

*Sensitivity analysis of with further diagnostic confirmation*

In addition to the primary analysis based on ICD-code defined pancreatic cancer, a sensitivity analysis was performed using a large language model (LLM)-based diagnostic verification pipeline. Inputs encompassed both structured data tables as well as radiology reports, surgical operative and procedure notes, pathology and surgical specimen reports, oncology and physician progress notes, and discharge summaries were evaluated. Clinical documents were then segmented into chunks which retained original wording and clinical semantics (e.g., coherent blocks such as assessment blocks, imaging impressions, operative summaries) and ensured no modification of clinical meaning. Diagnosis confirmation was defined as a clinical extraction task with specific prompting logic as described below. Prompts included explicit inclusion and exclusion criteria, strict handling of uncertainty language, domain specific guardrails, and strict temporal logic. Subsequent verifications were performed on JSON outputs using simple rule-based parses.

After task-scoped context retrieval, downstream inference used a local LLM on distributed GPU clusters. We used GPT-OSS-20B constrained as standard to a temperature of 0.70, nucleus sampling value of 0.9 and Top-K value of 10 with outputs to a JSON format. First a raw output was generated for each patient-task, which was subsequently parsed and validated. First a cohort based on ICD codes or a production-quality disease diagnosis model which uses three mentions in a BERT-based natural language processing (NLP) pipeline was run.

Prompts are documented below:

**System Prompt**

Your task is to read clinical notes and determine whether the patient has a CONFIRMED pancreatic

cancer diagnosis belonging to ANY of the following categories:

CATEGORY 1: Pancreatic ductal adenocarcinoma (PDAC)

• Pancreatic ductal adenocarcinoma (PDAC)

• Pancreatic adenocarcinoma

CATEGORY 2: Intraductal Papillary Mucinous Carcinoma (IPMC)

• Carcinoma arising from IPMN

• Invasive carcinoma associated with IPMN

• PDAC derived from IPMN

CATEGORY 3: Other exocrine pancreatic carcinomas

• Adenosquamous carcinoma

• Acinar cell carcinoma

• Mixed neuroendocrine–acinar carcinoma

• Colloid carcinoma

• Hepatoid carcinoma

• Pancreatoblastoma

CRITICAL RULE:

Do NOT classify as diagnosis if uncertainty language is present:

“concern for”, “possible”, “probable”, “likely”, “suspected”,

“rule out”, “query”, or differential diagnosis phrasing.

Only explicit CONFIRMED diagnoses count.

DATE FORMAT RULE:

Diagnosis date must be converted to YYYY-MM-DD regardless of original format.

Also extract the exact excerpt that explicitly confirms the diagnosis.

**User Prompt**From the provided clinical text, determine the following:

1. Is the patient a confirmed case of ANY of the following pancreatic cancer subtypes? (Yes/No)

- PDAC

- Pancreatic adenocarcinoma

- Intraductal Papillary Mucinous Carcinoma (IPMC)

- IPMN with associated invasive carcinoma

- Adenosquamous carcinoma

- Acinar cell carcinoma

- Mixed neuroendocrine–acinar carcinoma

- Colloid carcinoma

- Hepatoid carcinoma

- Pancreatoblastoma

2. If yes:

- Extract diagnosis_date (converted to YYYY-MM-DD)

- Extract excerpt_evidence (exact confirming text)

Return ONLY this JSON structure:

```
{

"is_pancreatic_cancer_confirmed": "Yes/No",

"diagnosis_date": "",

"excerpt_evidence": ""

}
```

RULES:

| |
|---|
| - YES only if the text explicitly confirms a listed subtype.<br>- If uncertainty language is present, return "No".<br>- If only “pancreatic cancer” is mentioned without subtype, return "No".<br>- diagnosis_date must be normalized to YYYY-MM-DD.<br>- excerpt_evidence must be the exact confirming phrase.<br>- Return ONLY valid JSON with no explanations. |

*Time series data representation*

In this study, we aim to determine the risk of pancreatic cancer using longitudinal diagnostic and blood-based biomarkers. A patient's diagnosis or blood-based biomarkers histories consist of a sequence of irregularly spaced visits, each containing a variable number of ICD diagnosis codes or blood tests.

To handle the irregularly sampled time series data, we first introduced a time-bucketed frequency encoding paradigm to translate irregular time series for diagnosis and blood test data per patient into structured matrices $\mathbf{X}_{\mathrm{C}} \in \mathbb{R}^{\mathrm{T}\times\mathrm{C}}$ and $\mathbf{X}_{\mathrm{L}} \in \mathbb{R}^{\mathrm{T}\times\mathrm{L}}$, where $\mathrm{C}$ and $\mathrm{L}$ is the total number of diagnosis codes ($\mathrm{C} = 2696$) and blood tests ($\mathrm{L} = 31$) observed in the entire dataset, respectively (corresponding to the vocabulary size in natural language processing), and $\mathrm{T}$ is the total number of buckets with each bucket represents a duration of $\tau = 30$ days. For the diagnosis encoding $\mathbf{X}_{\mathrm{C}}$, the value of element $x_{\mathrm{C}}^{(c,t)}$ corresponds to the occurrence frequency of the ICD code $c$ presented in bucket $t$. Meanwhile, the element value $x_{\mathrm{L}}^{(l,t)}$ of $\mathbf{X}_{\mathrm{L}}$ equals the average measurement of blood test item $l$ over time bucket $t$.

The clinical utility in grouping visits into buckets is that diagnostic information and blood-based measurements obtained within a short period, such as a month, can serve as a holistic snapshot of the patient's status at that moment when evaluating pancreatic cancer risk, rather than exhibiting separate temporal or causal dependencies.

Additionally, averaging the same blood tests within a bucket would yield a better estimate of the true value, leading to a more accurate representation of temporal trends. As such, this design would streamline data representation and modeling complexity.

**Figure 1A** shows examples of the encoding matrices for diagnosis codes ($\mathbf{X}_\mathrm{C}$) and blood-based tests ($\mathbf{X}_\mathrm{L}$), respectively, for a pancreatic cancer case and a control spanning 20 years prior to diagnosis. Note that, for better visualization in this case, diagnosis codes (blue matrix) were grouped into ICD chapters (x-axis), and the time bucket duration was set to 6 months (y-axis).

*Model*

**Figure 1B** demonstrates the overall architecture of the proposed pancreatic cancer risk detection model ('PRE-diagnostic Pancreatic Cancer Risk MODel'; PREMOD). Following the sequential modeling paradigm, the model was developed based on the Transformer encoder architecture,[47] which comprises four components: input encoder, feature extractor, feature aggregator, and classifier.

First, the row-wise concatenated encoding matrix of diagnosis codes and blood tests, $\mathbf{X} \in \mathbb{R}^{\mathrm{T}\times\mathrm{D}}$ $(\mathrm{D} = \mathrm{C} + \mathrm{L})$, was linearly projected into a lower-dimensional feature space $\mathbf{X}_\mathrm{I} \in \mathbb{R}^{\mathrm{T}\times d_\mathrm{model}}$, where $d_\mathrm{model}$ represents the lower feature space dimensionality. Then, a sinusoidal temporal encoding $\mathbf{P} \in \mathbb{R}^{\mathrm{T}\times d_\mathrm{model}}$, generated using the positional encoding method introduced in the original Transformer model,[47] was element-wise added to $\mathbf{X}_\mathrm{I}$ to incorporate temporal order of the time buckets. Next, a multi-head self-attention module (with $l$ layers and $h$ attention heads) was employed to extract features from all buckets, resulting in a feature matrix $\mathbf{Z} = (\boldsymbol{z}_0, \dots, \boldsymbol{z}_t, \dots, \boldsymbol{z}_{\mathrm{T}-1}) \in \mathbb{R}^{\mathrm{T}\times d_\mathrm{model}}$, where $\boldsymbol{z}_t \in \mathbb{R}^{d_\mathrm{model}}$ is the feature vector for bucket $t$.

We further introduced a multi-head self-attention aggregation module, inspired by the additive attention mechanism,[48] to pool extracted features from different time buckets into context heads: $\boldsymbol{c}_g = \boldsymbol{a}_g^\mathrm{T}\mathbf{Z} \in \mathbb{R}^{d_\mathrm{model}}$, where $g$ denotes the feature aggregation head and $\boldsymbol{a}_g = \sigma(MLP_g(\mathbf{Z})) \in \mathbb{R}^\mathrm{T}$ is the learnable feature weights (attentions) for the head $g$ obtained from feeding the feature matrix $\mathbf{Z}$ into a multi-layer perceptron model $MLP_g(\cdot)$ followed by a Softmax normalization function $\sigma(\cdot)$. The concatenated context heads were then linearly transformed to the context vector $\mathbf{c} \in \mathbb{R}^{d_{\text{model}}}$, which was subsequently used by the downstream classifier module (a two hidden layer multilayer perceptron) to detect the pancreatic cancer risk, and the classification result as shown in **Figure 1C**.

*Training and evaluation*

To better use the data while ensuring an objective assessment, we employed a leave-one-site-out (LOSO) strategy for model development and performance evaluation. Specifically, data from three different sites of the Mayo Clinic system, namely Mayo Clinic Health System (MCHS), Mayo Clinic Arizona, and Mayo Clinic Florida, were iteratively used as the independent testing set. Meanwhile, 10-fold cross-validation was applied to the remaining data, namely the development set, for model training, validation, and hyperparameter tuning. Note that data associated with the Mayo Clinic Rochester were used for model development in all LOSO iterations, as it accounts for the largest proportion of the data (~46%) across all sites. As a result, 10 trained models were obtained in each LOSO iteration and were evaluated on the corresponding test set. The mean and 95% confidence interval (CI) of the Area Under the Receiver Operating Characteristic curve (AUROC) across all three test sets were calculated and reported.

As investigated by Placido et al., we used the complete time-series data for model training, i.e., without an exclusion intervals.[15] Then, the trained model was evaluated for predicting pancreatic cancer risk with 1-, 2-, and 3-year lead times (i.e., prediction intervals), where data were excluded from respective lead-time periods during model testing. This approach is justified in that testing is performed in the real-world deployment scenario with temporal masking offering differing masks to account for varying time courses of clinical work up and any associated imperfections to the documented date of diagnosis. With this approach we expect patients amidst work-up for pancreatic cancer to be present in the training corpus and therefore should accurately reflect a higher risk, as compared to the pre-work-up population where the learned signals play a more significant role. The robustness of this approach was also confirmed in the methods of Placido et al.[3]

During modeling training, we employed multiple methods to reduce the impact of data imbalance. First, a down-sampling strategy with a $1:10$ case-to-control ratio was used, which helped the model learn more effectively from positive cases and further improved training efficiency. Additionally, we adopted Focal Loss as the training objective for the model to focus on minority classes. The loss parameters were set to $\gamma = 2.0$ and $\alpha = 0.25$ as suggested in Lin et al.[49] Then, the performance was assessed on test sets with original data distributions. These methods are known to skew model calibration with inflated risk estimates.[50,51] To account for this, a custom Bayesian recalibration was performed on model outputs that only require an assumption of the pancreatic cancer prevalence (explained below).[52]

All hyperparameters were determined via grid search based on the average performance on the development sets (**Supplementary Table 4**). For model design, we evaluated bucket sizes $\tau \in \{7, 30, 90\}$ and selected $\tau = 30$ (monthly bucket). The number of multi-head self-attention layers and heads of the Transformer encoder component were set to $2$ and $16$, respectively, which were chosen from the sets $\{1, 2, 4, 8\}$ and $\{4, 8, 16, 32\}$. The dimension of the feature embedding space $d_{\text{model}}$ was $512$, selected from $\{256, 512, 1024\}$. The number of feature aggregation heads $g$ was set to $2$, with a search over $\{1, 2, 3, 4, [\text{CLS}]\}$, where $[\text{CLS}]$ presents using a dedicated learnable token for downstream classification instead of the attention-based feature aggregation. During modeling training, the initial learning rate was set to $2 \times 10^{-4}$, selected from $\{10^{-4},\ 2 \times 10^{-4},\ 3 \times 10^{-4},\ 5 \times 10^{-4}\}$, and reduced by a factor of 5 after two stagnant epochs measured by validation loss. The batch size was set to $64$, chosen from $\{16, 32, 64, 128\}$. The $1{:}10$ down-sampling ratio in the training was selected from $\{1{:}1,\ 1{:}5,\ 1{:}10,\ 1{:}20,\ \text{None}\}$, where $\text{None}$ indicates no resampling was performed.

To assess the model's diagnostic effectiveness, we employed the metric of diagnostic odds ratio (DOR), a single summative measure of a model's discriminative power.[20] Specifically, DOR is defined as the ratio of the odds of true-positive diagnosis in diseased cases to the odds of false-positive diagnosis in non-diseased individuals. It is formulated as

$$\text{DOR} = \frac{\text{TP}/\text{FP}}{\text{FP}/\text{TN}} = \frac{\text{TP} \cdot \text{TN}}{\text{FP} \cdot \text{FN}} \qquad (1)$$

where $\mathrm{TP}$, $\mathrm{FP}$, $\mathrm{TN}$, and $\mathrm{FN}$ represent the model's true-positive, false-positive, true-negative, and false-negative predictions. DOR ranges from 0 to infinity, with higher values indicating better diagnostic performance. When $\mathrm{DOR} = 1$, it suggests the model provides no diagnostic information, while $\mathrm{DOR} > 1$ indicating a better performance. Since $\mathrm{DOR}$ relies on the ratio of odds, it is independent of disease prevalence and often remains relatively stable to the decision threshold.[20]

Furthermore, to better understand the model's decision, we applied the SHapley Additive exPlanations (SHAP) analysis to predictions in the test set.[53] The resulting matrix of SHAP values follow the same dimensionality of the input data, denoted $\mathbf{S} \in \mathbb{R}^{\mathrm{N}\times\mathrm{T}\times\mathrm{D}}$ where $\mathrm{N}$, $\mathrm{T}$, and $\mathrm{D}$ represent the sample, time, and feature dimensions, respectively. By aggregating over the time dimension, we calculated the mean absolute SHAP values across all patients, yielding a feature contribution vector, to obtain the top 20 contributing features (ICD codes and blood tests). Similarly, mean absolute SHAP values across the feature and sample dimensions were calculated to demonstrate the time-wise contribution to the model's predictions over the patient's clinical trajectory.

*Bayesian prevalence-aware recalibration*

To counteract a learned inflated prior from our resampled training set, where the observed prevalence of pancreatic cancer is higher than in testing and real-world populations, we introduced a Bayesian post-hoc recalibration strategy that accounts for the prior probability (i.e., expected prevalence) of pancreatic cancer in the target dataset.[52]

Specifically, our Transformer model estimates the probability that a patient has pancreatic cancer ($y = 1$) given the input features $\mathbf{X} \in \mathbb{R}^{\mathrm{T}\times\mathrm{D}}$, i.e., $P(y = 1 \mid \mathbf{X})$. Per Bayes' Theorem, the posterior odds of pancreatic cancer, i.e., the cancer to no-cancer ratio produced by the model, are the product of the prior belief of the disease prevalence in the dataset (namely prior odds) and the likelihood ratio ($LR$)

$$\frac{P(y=1 \mid \mathbf{X})}{P(y=0 \mid \mathbf{X})} = \frac{P(y=1)}{P(y=0)} \times \frac{P(\mathbf{X} \mid y=1)}{P(\mathbf{X} \mid y=0)} \qquad (2)$$

where the likelihood ratio represents the diagnostic evidence learned by the model, a fixed property of the model that is invariant across different clinical environments. In contrast, the prior odds are a property of the specific dataset or data sampling strategy. Based on Equation (2), the uncalibrated posterior odds (denoted $\Omega_u$) and calibrated posterior odds (denoted $\Omega_c$) can be respectively derived as

$$\Omega_u = LR \times \pi_{src} \qquad (3)$$

$$\Omega_c = LR \times \pi_{tar} \qquad (4)$$

where $\pi_{src}$ is the prior odds of pancreatic cancer in the training distribution ($\pi_{src} = 1/10$ in our case) and $\pi_{tar}$ is the prior odds in the target clinical population. Dividing Equation (4) by (3), we define the relationship between calibrated and uncalibrated posterior odds states as

$$\frac{\Omega_c}{\Omega_u} = \frac{LR \times \pi_{tar}}{LR \times \pi_{src}} \qquad (5)$$

Further canceling out the likelihood ratio term, followed by taking the logarithm on both sides, the calibrated odds in the log-space can be formulated as

$$\log \Omega_c = \log \Omega_u + \log \pi_{tar} - \log \pi_{src} \qquad (6)$$

In the binary classification framework using a sigmoid activation function ($\sigma = \frac{1}{1+e^{-z}}$), the model's raw uncalibrated logit output ($z_u$) is mathematically equivalent to the log-odds of the disease, which is given by

$$z_u = \sigma^{-1}(p) = \log\left(\frac{p}{1-p}\right) = \log(\frac{P(y=1 \mid \mathbf{X})}{P(y=0 \mid \mathbf{X})}) = \log \Omega_u \qquad (7)$$

where $p = P(y = 1 \mid \mathbf{X})$ is the model's estimated probability of pancreatic cancer. Therefore, we transition from the log-odds in Equation (6) to the logit space to derive the logit adjustment expression

$$z_c = z_u + \Delta \qquad (8)$$

where $z_c$ and $z_u$ are the calibrated and raw logit produced by the model, respectively, and $\Delta = \log \pi_{tar} - \log \pi_{src}$ represents the adjustment term to the model's logit output. As the adjustment is a linear shift in the logit space, it allows the model to remain discriminatively robust (i.e., preserving AUROC) while providing context-specific probabilities suitable for diverse clinical environments. This allows for the model to be deployed in diverse settings where the pancreas cancer prevalence can be updated based on the latest epidemiological insights. To evaluate the calibration efficacy, we compared the Brier score, the slope and intercept of the linear regression lines fitted from the calibration curves, and the Expected Calibration Error.[54]

This approach is advantageous in that the prior odds can be modified in reference to different target populations ($\pi_{tar}$) that the model is intended to be deployed in. This calibration framework allows the model to be recalibrated for different clinical contexts, enabling adaptive and broader clinical applications. By comparison, other calibration techniques, such as isotonic regression[55] and temperature scaling,[56] would require additional model training or retuning based on a labeled calibration set with the same data distribution as the target population.

*Simulation of real-world continuous risk prediction*

To evaluate the clinical utility of our Transformer-based model in a real-world setting, we simulated a continuous risk prediction framework using a temporal sliding window approach. As each patient's diagnosis and laboratory data was encoded as a matrix $\mathbf{X} \in \mathbb{R}^{\mathrm{T} \times \mathrm{D}}$, we applied a fixed-length ($\mathrm{T}$) sliding window along the temporal dimension $\mathrm{T}$ with a 1-month step size. Data points falling outside the 20-year pre-diagnosis window were zero-padded to maintain consistent input dimensionality for the Transformer architecture. As such, this process transformed each patient's clinical trajectory into a series of matrices with the same shape, effectively capturing the temporal evolution of their clinical profile.

Consequently, the trained model generated a continuous series of probability scores for each sliding window, which were then used to construct longitudinal risk curves. This allowed us to visualize the dynamic change in predicted risk as the model integrated progressively more recent clinical data. To better understand the evolution of risk over time, we generated an individual risk trajectory for each patient to observe patient-specific signal emergence (**Figure 1D**), as well as the aggregated risk trends across all patients to identify the standard lead time for risk elevation (**Figure 3**).

*Modeling population-level screening feasibility*

The extreme rarity of pancreatic cancer makes population-level screening prohibitive. Lung cancer screening uses a sensitive demographic cut-off (age 50-80 years and 20 pack year history) to enrich a high prevalence cohort for low dose CT screening to act as a diagnostic filter ahead of confirmatory invasive lung biopsy.[57] Breast cancer screening is also made feasible because age (40-74 years) sufficiently enriches a high prevalence cohort for mammographic screening prior to confirmatory biopsy.[58]

To achieve a comparable screening pipeline in pancreatic cancer, we operationalize our model as a high-sensitivity, population-level digital screen amongst those aged 50-74 years, where 90% of pancreatic cancers occur;[27] followed by an AI-enabled non-invasive CT screen to-enrich a cohort for endoscopic ultrasound (EUS) and biopsy confirmation. Here, REDMOD serves as the corollary to low-dose CT in lung cancer screening or mammography in breast cancer screening.[26]

The pancreatic cancer prevalence in a population of 50–74-year-olds was determined based on age-standardized rates (ASR) from the SEER database and population sizes from census data (63.2 million individuals aged 50-64 and 31.2 million individuals aged 65-74):

$$ASR = \sum_{i=1}^{N} r_i \times w_i \qquad (9)$$

where $r_i$ is the age-specific rate for age group $i$, $w_i$ is the weight of age group $i$ derived from the standard reference population, and $N$ is the number of age groups.

To evaluate the feasibility of this approach to population-level screening, we modeled a multi-stage Bayesian pipeline:

- Stage 1: High-sensitivity digital screen using our model (PREMOD);
- Stage 2: High-specificity filter using non-invasive CT with REDMOD AI algorithm (externally validated to detect visually occult pancreatic cancer with 475 day lead-time; 73% sensitivity and 81% specificity);[26]
- Stage 3: Diagnostic confirmation via EUS and biopsy (90.8% sensitivity and 94% specificity from meta-analyses).[31,32]

For each screening stage $s \in \{1,2,3\}$, the positive cases identified by the model (i.e., true positives + false positives) were calculated from

$$N_s^+ = [\mathrm{Sensitivity_s} \times P_s + (1 - \mathrm{Specificity_s}) \times (1 - P_s)] \times N_s \qquad (10)$$

where $\mathrm{Sensitivity_s}$ (true positive rate) and $\mathrm{Specificity_s}$ (true negative rate) were from the model's detections at the current stage, $P_s$ and $N_s$ represent the prevalence and population that were derived from the previous stage.

In our analyses, the population prevalence of pancreatic cancer for digital screening, was taken at 33.2 per 100,000 (age-standardized rate amongst 50-74 year olds, per SEER), i.e., $P_1 = 0.000332$ and $N_1 = 100{,}000$.[59] Performance metrics for Stage 1 were taken from the MCHS test set (community-based centers) to best reflect wider national adoption (vs the other tertiary and quaternary test hospitals). To operationalize our model, we tuned the classification threshold to the expected prevalence of the deployment setting (matching the 3.3% prevalence in our testing data), yielding a "high-sensitivity" model with $\text{Senstivity}_1 = 95.3\%$ and $\text{Specificity}_1 = 55.2\%$.

For Stages 2 and 3, the prevalence $P_2$ and $P_3$ were given by the corresponding post-test probability from the previous stage, derived from the Bayesian calculation as

$$P_{s\in\{2,3\}} = \frac{\text{Sensitivity}_s \times P_{s-1}}{(\text{Sensitivity}_s \times P_{s-1}) + (1 - \text{Specificity}_s) \times (1 - P_{s-1})} \qquad (11)$$

where $\text{Senstivity}_s$ served as the likelihood and the prevalence $P_{s-1}$ was the prior. Likewise, the screening population in stage $s \in \{1,2\}$ was also defined by the previous model's detection, i.e., $N_s := N_{s-1}^{+}$. In the analyses, we adopted the published diagnostic performance of $\text{Sensitivity}_2 = 73.0\%$ and $\text{Specificity}_2 = 81.0\%$ for REDMOD in Stage 2 and $\text{Sensitivity}_3 = 90.8\%$ and $\text{Specificity}_3 = 94.0\%$ for the EUS screening in Stage 3.

Therefore, the Number Needed to Screen (NNS) to detect one pancreatic cancer in the multi-stage screening pipeline was calculated as

$$NNS = \frac{1}{P_3 \times \text{Sensitivity}_3}. \qquad (12)$$

Moreover, the number of true positive cancer detections $(\text{TP})$ from the screening pipeline was calculated as $\text{TP} = P_3 \times N_3^{+}$.

The detection efficiency was quantified as the fold reduction in $NNS$ compared to direct EUS screening of an unselected population $NNS_{base} = 1/(P_1 \times \text{Sensitivity}_3)$. We also compare screening cohort enrichment against the END-PAC model applied to new-onset diabetes populations using the same framework (see **Supplementary Results**).[60]

This simulation is intended only to model what a possible digitally enriched high-risk cohort, with downstream radiologic and biopsy-based screening interventions would look like and is not intended to replace the need for prospective evaluation. We also use the literature to inform the performance characteristics of downstream tests where population-level performance is not always reported (e.g., for REDMOD). It is therefore, acknowledged that further evaluation in a population-based cohort is necessary and now justified by the presented results.

## Supplementary Results

*Differential contribution of blood-test biomarkers and medical histories*

For the 1-year prediction lead time, the across-sites mean AUROC for models trained separately on diagnosis and lab test histories was 0.829 [95% CI: 0.823—0.835] and 0.638 [95% CI: 0.623—0.652], respectively, both of which were inferior to using the combined data (**Supplementary Figure 3**). This may suggest that the two different data types would contain complementary information for pancreatic cancer detection.

*Model recalibration*

With recalibration, the mean expected calibration error, calibration slope, and intercept were $0.037, 1.075, -0.077$ for 1-year prediction lead time, $0.043, 1.034, -0.1$ for 2-year prediction lead time, and $0.051, 0.991, -0.118$ for 3-year prediction lead time. By comparison, the mean ECEs were over $3$x higher without recalibration, at $0.127, 0.844, -0.181$ for 1-year prediction lead time, $0.142, 0.804, -0.172$, for 2-year prediction lead time, and $0.161, 0.723, -0.159$ for 3-year prediction lead times.

*Simulated digital enrichment toward a first screening paradigm for pancreatic cancer*

To make pancreatic cancer screening possible despite its challenging low-prevalence, we set our model to a high-sensitivity mode by tuning the classification threshold to the expected pancreatic cancer prevalence at deployment (sensitivity of 95.3%, at classification threshold of 3.3%, based on test-set prevalence,[32,33] with 1-year lead-time; **Supplementary Table 3**), and then utilize a high-specificity modality, a non-invasive computed tomography (CT) scan with Radiomics-based Early Detection MODel (REDMOD) analysis (rationale explained in **Methods; '***Modeling population-level screening feasibility***')**,[30,34] with individuals screening positive on both tests proceeding to endoscopic ultrasound (EUS) biopsy confirmation. When operationalizing this pipeline in a setting with an age-standardized population prevalence of sporadic pancreatic cancer of 33.2 per 100,000 among those aged 50-74 (as >90% of pancreatic cancers occur in those aged >55),[35,36,12,37] our threshold-tuned model will identify 44,816 individuals in a population of 100,000 for radiomic screening. The REDMOD model with an external validation sensitivity of 73% and specificity of 81% at an optimized threshold,[38,39] will then identify 8,532 patients for EUS screening which, with a conservative 90.8% sensitivity and 94% specificity,[17,30] will identify 21 cancers; translating to a number needed to screen to detect one pancreatic cancer of 406. This pipeline improves screening efficiency by 8.2x compared to if EUS biopsy was used in an unenriched population (NNS of 3317 to detect one pancreatic cancer) and represents the state of the art.

Using this pipeline achieves a PPV of 3.9% in the general population, which exceeds the literature-reported 2-3.6% PPV of the END-PAC in a pre-enriched new-onset diabetes cohorts, which notably misses many pancreatic cancers and the modelled END-PAC PPV of 0.6% when applied to an unenriched general 50-74 year-old population (acknowledging that END-PAC was developed for use within glycemically defined new-onset diabetes population).[19] Our pipeline also supersedes the PPV of 3.7% reported for low dose CT screening in 50-80 year-olds with a 20-pack year smoking history for T0 lung cancer confirmation.[40]

Using a higher threshold such as the Youden's J statistic maximized threshold of 6.6% in the community setting for a 1-year prediction lead time, in the same population would identify 30,418 high-risk individuals. Using REDMOD, 5,794 individuals for EUS, of whom 25 will have cancer (NNS 876, 3.8x improved screening efficiency).

*Comparison to END-PAC in New-Onset Diabetes Cohort*

Importantly, models based on new-onset diabetes are not suitable for population-based screening as >60% of sporadic pancreatic cancers occur without measurable new onset diabetes.[12,39,40] Hence approaches relying on diagnosis of new-onset diabetes,[12] only capture 34%-40% of patients with pancreatic cancer.[13,14] In addition, prior meta-analyses show that an END-PAC score ≥3 applied in a new-onset diabetes cohort showed a pooled sensitivity of 55.8% and specificity of 82.0%.[59]

Hence, if we assume an optimistic estimate of 40% of our proposed 100,000 50-74 year olds put forward for screening had new-onset diabetes, an END-PAC score of ≥3 identifies 18,005 as high-risk, which if you directed to EUS biopsy, would detect 6.7 cancers with a number needed to screen of 2,676 to detect 1 cancer (3.1x fold improvement to EUS biopsy in an unenriched cohort). This translates to an overall screening PPV of 0.6%.

**Supplemental Table 1:** Demographic summary of cohort, stratified by site

| Site | All (N=183,098) | Pancreatic cancer cases (N=6,017) | Non-cancer controls (N=177,081) |
|---|---|---|---|
| **ALL SITES** | | | |
| **Diagnosis Age** (median [IQR]) | 75 [66, 82] | 73 [65, 79] | 75 [66, 82] |
| **Gender** | | | |
| Female | 81833 (44.7%) | 2830 (47.0%) | 79003 (44.6%) |
| **Race** | | | |
| White | 170988 (93.4%) | 5707 (94.9%) | 165281 (93.3%) |
| Black | 3487 (1.9%) | 92 (1.5%) | 3395 (1.9%) |
| Asian | 2353 (1.3%) | 63 (1.0%) | 2290 (1.3%) |
| Native American/Pacific Islander | 761 (0.4%) | 24 (0.4%) | 737 (0.4%) |
| Other | 5509 (3.0%) | 131 (2.2%) | 5378 (3.0%) |
| **ROCHESTER** | **(N=84,350)** | **(N=3,185)** | **(N=81,165)** |
| **Diagnosis Age** (median [IQR]) | 75 [66, 82] | 72 [64, 79] | 75 [66, 82] |
| **Gender** | | | |
| Female | 36131 (42.8%) | 1451 (45.6%) | 34680 (42.7%) |
| **Race** | | | |
| White | 78013 (92.5%) | 3010 (94.5%) | 75003 (92.4%) |
| Black | 1047 (1.2%) | 27 (0.8%) | 1020 (1.3%) |
| Asian | 1028 (1.2%) | 31 (1.0%) | 997 (1.2%) |
| Native American/Pacific Islander | 305 (0.4%) | 10 (0.3%) | 295 (0.4%) |
| Other | 3957 (4.7%) | 107 (3.4%) | 3850 (4.7%) |
| **MCHS** | **(N=52,218)** | **(N=1,451)** | **(N=50,767)** |
| **Diagnosis Age** (median [IQR]) | 74 [65, 83] | 73 [65, 80] | 74 [65, 83] |
| **Gender** | | | |
| Female | 24965 (47.8%) | 717 (49.4%) | 24248 (47.8%) |
| **Race** | | | |
| White | 50704 (97.1%) | 1426 (98.3%) | 49278 (97.1%) |
| Black | 400 (0.8%) | 6 (0.4%) | 394 (0.8%) |
| Asian | 369 (0.7%) | 6 (0.4%) | 363 (0.7%) |
| Native American/Pacific Islander | 155 (0.3%) | 2 (0.1%) | 153 (0.3%) |
| Other | 590 (1.1%) | 11 (0.8%) | 579 (1.1%) |
| **ARIZONA** | **(N=23,588)** | **(N=724)** | **(N=22,864)** |
| **Diagnosis Age** (median [IQR]) | 75 [67, 82] | 73 [67, 80] | 75 [67, 82] |
| **Gender** | | | |
| Female | 10056 (42.6%) | 331 (45.7%) | 9725 (42.5%) |
| **Race** | | | |
| White | 21948 (93.0%) | 686 (94.8%) | 21262 (93.0%) |
| Black | 484 (2.1%) | 11 (1.5%) | 473 (2.1%) |
| Asian | 483 (2.0%) | 13 (1.8%) | 470 (2.1%) |
| Native American/Pacific Islander | 226 (1.0%) | 11 (1.5%) | 215 (0.9%) |
| Other | 447 (1.9%) | 3 (0.4%) | 444 (1.9%) |
| **FLORIDA** | **(N=22,942)** | **(N=657)** | **(N=22,285)** |
| **Diagnosis Age** (median [IQR]) | 74 [66, 81] | 73 [65, 79] | 74 [66, 81] |
| **Gender** | | | |

| | | | | |
|---|---|---|---|---|
| | Female | 10681 (46.6%) | 331 (50.4%) | 10350 (46.4%) |
| **Race** | | | | |
| | White | 20323 (88.6%) | 585 (89.0%) | 19738 (88.6%) |
| | Black | 1556 (6.8%) | 48 (7.3%) | 1508 (6.7%) |
| | Asian | 473 (2.1%) | 13 (2.0%) | 460 (2.1%) |
| | Native American/Pacific Islander | 75 (0.3%) | 1 (0.2%) | 74 (0.3%) |
| | Other | 515 (2.2%) | 10 (1.5%) | 505 (2.3%) |

**Supplemental Table 2:** Diagnostic confirmation sensitivity analysis

| Testing cohort | Testing Site | AUROC |
|---|---|---|
| **LLM-confirmed** (1353: 95889) | MCHS (687: 50756) | 0.842 |
| | Arizona (333: 22853) | 0.815 |
| | Florida (333: 22280) | 0.778 |
| | Overall (1353: 95889) | 0.812 |
| **Original ICD-based** (2832: 95916) | MCHS (1451: 50767) | 0.872 |
| | Arizona (724: 22864) | 0.830 |
| | Florida (657: 22285) | 0.810 |
| | Overall (2832: 95916) | 0.837 |

**Supplementary Table 3:** Performance metrics across various thresholds

| Prediction lead time: 1 year | | | | | | | | |
|---|---|---|---|---|---|---|---|---|
| **Site** | **Threshold Criteria** | **Threshold** | **Specificity** | **Sensitivity** | **PPV** | **NPV** | **Youden's J** | **DOR** |
| MCHS | Sensitivity = 0.8 | 0.079 | 0.748 | 0.800 | 0.084 | 0.993 | 0.550 | 12.4 |
| | Specificity = 0.8 | 0.092 | 0.800 | 0.734 | 0.097 | 0.991 | 0.535 | 11.5 |
| | Sensitivity = Specificity | 0.085 | 0.770 | 0.770 | 0.090 | 0.992 | 0.547 | 12.1 |
| | Max Youden's J | 0.066 | 0.696 | 0.859 | 0.075 | 0.994 | 0.554 | 14.4 |
| | Threshold = 0.033 (prior prevalence) | 0.033 | 0.552 | 0.953 | 0.058 | 0.998 | 0.506 | 27.1 |
| | Threshold = 0.5 (default) | 0.5 | 0.997 | 0.256 | 0.707 | 0.979 | 0.253 | 120.7 |
| Arizona | Sensitivity = 0.8 | 0.073 | 0.651 | 0.800 | 0.068 | 0.990 | 0.451 | 7.67 |
| | Specificity = 0.8 | 0.101 | 0.800 | 0.658 | 0.097 | 0.987 | 0.460 | 7.94 |
| | Sensitivity = Specificity | 0.088 | 0.730 | 0.730 | 0.080 | 0.989 | 0.461 | 7.49 |
| | Max Youden's J | 0.095 | 0.767 | 0.692 | 0.088 | 0.988 | 0.462 | 7.64 |
| | Threshold = 0.033 (prior prevalence) | 0.033 | 0.463 | 0.927 | 0.052 | 0.995 | 0.389 | 11.7 |
| | Threshold = 0.5 (default) | 0.5 | 0.999 | 0.172 | 0.786 | 0.974 | 0.171 | 145 |
| Florida | Sensitivity = 0.8 | 0.068 | 0.599 | 0.800 | 0.058 | 0.991 | 0.402 | 6.67 |
| | Specificity = 0.8 | 0.096 | 0.800 | 0.638 | 0.087 | 0.987 | 0.435 | 7.18 |
| | Sensitivity = Specificity | 0.083 | 0.712 | 0.712 | 0.070 | 0.988 | 0.424 | 6.43 |
| | Max Youden's J | 0.101 | 0.826 | 0.612 | 0.096 | 0.986 | 0.438 | 7.69 |
| | Threshold = 0.033 (prior prevalence) | 0.033 | 0.293 | 0.955 | 0.034 | 0.996 | 0.247 | 15.7 |
| | Threshold = 0.5 (default) | 0.5 | 0.997 | 0.202 | 0.688 | 0.977 | 0.199 | 103 |
| Overall | Sensitivity = 0.8 | 0.073 | 0.671 | 0.800 | 0.070 | 0.991 | 0.471 | 9.06 |
| | Specificity = 0.8 | 0.097 | 0.800 | 0.673 | 0.094 | 0.988 | 0.475 | 8.78 |
| | Sensitivity = Specificity | 0.085 | 0.740 | 0.740 | 0.080 | 0.990 | 0.478 | 8.68 |
| | Max Youden's J | 0.086 | 0.743 | 0.735 | 0.081 | 0.990 | 0.478 | 8.67 |
| | Threshold = 0.033 (Prior prevalence) | 0.033 | 0.436 | 0.945 | 0.050 | 0.996 | 0.381 | 18.2 |
| | Threshold = 0.5 (default) | 0.5 | 0.996 | 0.210 | 0.727 | 0.977 | 0.208 | 123 |
| **Prediction lead time: 2 years** | | | | | | | | |
| **Site** | **Threshold Criteria** | **Threshold** | **Specificity** | **Sensitivity** | **PPV** | **NPV** | **Youden's J** | **DOR** |
| MCHS | Sensitivity = 0.8 | 0.082 | 0.702 | 0.800 | 0.072 | 0.992 | 0.503 | 9.62 |
| | Specificity = 0.8 | 0.104 | 0.800 | 0.668 | 0.087 | 0.988 | 0.467 | 8.06 |
| | Sensitivity = Specificity | 0.092 | 0.745 | 0.745 | 0.078 | 0.990 | 0.490 | 8.62 |
| | Max Youden's J | 0.068 | 0.641 | 0.872 | 0.078 | 0.994 | 0.513 | 12.4 |
| | Threshold = 0.033 (prior prevalence) | 0.033 | 0.477 | 0.968 | 0.051 | 0.998 | 0.444 | 30.4 |
| | Threshold = 0.5 (default) | 0.5 | 0.997 | 0.206 | 0.654 | 0.978 | 0.202 | 86.6 |
| Arizona | Sensitivity = 0.8 | 0.069 | 0.593 | 0.800 | 0.059 | 0.989 | 0.391 | 5.83 |
| | Specificity = 0.8 | 0.105 | 0.800 | 0.588 | 0.089 | 0.984 | 0.390 | 5.99 |
| | Sensitivity = Specificity | 0.086 | 0.701 | 0.701 | 0.070 | 0.987 | 0.404 | 5.65 |
| | Max Youden’s J | 0.086 | 0.701 | 0.701 | 0.070 | 0.987 | 0.404 | 5.65 |
| | Threshold = 0.033 (prior prevalence) | 0.033 | 0.377 | 0.938 | 0.046 | 0.995 | 0.315 | 10.4 |
| | Threshold = 0.5 (default) | 0.5 | 0.999 | 0.124 | 0.718 | 0.973 | 0.122 | 93.7 |
| Florida | Sensitivity = 0.8 | 0.065 | 0.501 | 0.800 | 0.046 | 0.989 | 0.304 | 4.32 |
| | Specificity = 0.8 | 0.099 | 0.800 | 0.543 | 0.074 | 0.983 | 0.340 | 4.76 |
| | Sensitivity = Specificity | 0.083 | 0.667 | 0.667 | 0.056 | 0.986 | 0.334 | 4.12 |
| | Max Youden’s J | 0.106 | 0.836 | 0.506 | 0.085 | 0.983 | 0.342 | 5.32 |
| | Threshold = 0.033 (prior prevalence) | 0.033 | 0.251 | 0.947 | 0.036 | 0.994 | 0.198 | 7.15 |
| | Threshold = 0.5 (default) | 0.5 | 0.997 | 0.158 | 0.596 | 0.976 | 0.154 | 61.5 |
| Overall | Sensitivity = 0.8 | 0.072 | 0.611 | 0.800 | 0.059 | 0.990 | 0.408 | 7.08 |
| | Specificity = 0.8 | 0.102 | 0.800 | 0.603 | 0.083 | 0.985 | 0.402 | 6.33 |
| | Sensitivity = Specificity | 0.086 | 0.706 | 0.706 | 0.068 | 0.988 | 0.413 | 6.31 |
| | Max Youden’s J | 0.086 | 0.706 | 0.706 | 0.068 | 0.988 | 0.413 | 6.31 |
| | Threshold = 0.033 (Prior prevalence) | 0.033 | 0.368 | 0.951 | 0.044 | 0.996 | 0.319 | 16.0 |
| | Threshold = 0.5 (default) | 0.5 | 0.997 | 0.162 | 0.656 | 0.976 | 0.160 | 80.6 |
| **Prediction lead time: 3 years** | | | | | | | | |

| Site | Threshold Criteria | Threshold | Specificity | Sensitivity | PPV | NPV | Youden's J | DOR |
|---|---|---|---|---|---|---|---|---|
| MCHS | Sensitivity = 0.8 | 0.083 | 0.642 | 0.800 | 0.061 | 0.991 | 0.442 | 7.32 |
| | Specificity = 0.8 | 0.116 | 0.800 | 0.575 | 0.077 | 0.985 | 0.375 | 5.48 |
| | Sensitivity = Specificity | 0.096 | 0.710 | 0.710 | 0.066 | 0.989 | 0.412 | 6.09 |
| | Max Youden's J | 0.073 | 0.593 | 0.854 | 0.057 | 0.993 | 0.448 | 8.73 |
| | Threshold = 0.033 (prior prevalence) | 0.033 | 0.390 | 0.970 | 0.044 | 0.998 | 0.360 | 22.0 |
| | Threshold = 0.5 (default) | 0.5 | 0.996 | 0.162 | 0.578 | 0.977 | 0.158 | 59.3 |
| Arizona | Sensitivity = 0.8 | 0.071 | 0.557 | 0.800 | 0.054 | 0.989 | 0.356 | 5.14 |
| | Specificity = 0.8 | 0.115 | 0.800 | 0.550 | 0.080 | 0.983 | 0.348 | 4.88 |
| | Sensitivity = Specificity | 0.093 | 0.681 | 0.681 | 0.064 | 0.986 | 0.362 | 4.60 |
| | Max Youden's J | 0.096 | 0.701 | 0.662 | 0.066 | 0.985 | 0.363 | 4.62 |
| | Threshold = 0.033 (prior prevalence) | 0.033 | 0.347 | 0.938 | 0.044 | 0.995 | 0.285 | 9.03 |
| | Threshold = 0.5 (default) | 0.5 | 0.998 | 0.109 | 0.641 | 0.973 | 0.107 | 64.3 |
| Florida | Sensitivity = 0.8 | 0.066 | 0.429 | 0.800 | 0.041 | 0.987 | 0.229 | 3.51 |
| | Specificity = 0.8 | 0.105 | 0.800 | 0.475 | 0.067 | 0.981 | 0.276 | 3.72 |
| | Sensitivity = Specificity | 0.086 | 0.633 | 0.633 | 0.050 | 0.984 | 0.267 | 3.19 |
| | Max Youden's J | 0.105 | 0.800 | 0.475 | 0.067 | 0.981 | 0.276 | 3.72 |
| | Threshold = 0.033 (prior prevalence) | 0.033 | 0.153 | 0.965 | 0.033 | 0.995 | 0.118 | 6.67 |
| | Threshold = 0.5 (default) | 0.5 | 0.997 | 0.122 | 0.549 | 0.975 | 0.118 | 49.1 |
| Overall | Sensitivity = 0.8 | 0.073 | 0.553 | 0.800 | 0.052 | 0.989 | 0.351 | 5.67 |
| | Specificity = 0.8 | 0.111 | 0.800 | 0.540 | 0.075 | 0.983 | 0.339 | 4.81 |
| | Sensitivity = Specificity | 0.091 | 0.677 | 0.677 | 0.060 | 0.986 | 0.354 | 4.77 |
| | Max Youden's J | 0.087 | 0.649 | 0.707 | 0.058 | 0.987 | 0.356 | 4.92 |
| | Threshold = 0.033 (Prior prevalence) | 0.033 | 0.297 | 0.958 | 0.040 | 0.996 | 0.254 | 12.6 |
| | Threshold = 0.5 (default) | 0.5 | 0.997 | 0.131 | 0.589 | 0.975 | 0.128 | 57.6 |

**PPV**: Positive Predictive Value; **NPV**: Negative Predictive Value; **Youden's J** = Sensitivity + Specificity - 1; **DOR**: Diagnosis Odds Ratio; **Prior prevalence**: Assumption of the disease prevalence in the targeted population.

**Supplementary Table 4**: Hyperparameter tuning within the development set

| Time bucket size in data encoding ($\tau$: days) | Feature extractor attention layers ($l$) | Feature extractor attention heads ($h$) | Feature dimension ($d_{\text{model}}$) | Feature aggregation attention heads ($g$) | Training down-sampling ratio | Initial learning rate | Training mini-batch size | Loss Function | AUROC |
|---|---|---|---|---|---|---|---|---|---|
| 30 | 2 | 16 | 512 | 2 | 1: 10 | 0.0002 | 64 | Focal | 0.874 |
| 7 | | | | | | | | | 0.838 |
| 90 | | | | | | | | | 0.835 |
| | 1 | | | | | | | | 0.864 |
| | 4 | | | | | | | | 0.863 |
| | 8 | | | | | | | | 0.664 |
| | | 4 | | | | | | | 0.867 |
| | | 8 | | | | | | | 0.869 |
| | | 32 | | | | | | | 0.863 |
| | | | 256 | | | | | | 0.833 |
| | | | 1024 | | | | | | 0.814 |
| | | | | 1 | | | | | 0.865 |
| | | | | 3 | | | | | 0.873 |
| | | | | 4 | | | | | 0.868 |
| | | | | [CLS] | | | | | 0.861 |
| | | | | | 1: 1 | | | | 0.829 |
| | | | | | 1: 5 | | | | 0.864 |
| | | | | | 1: 20 | | | | 0.789 |
| | | | | | None | | | | 0.841 |
| | | | | | | 0.0001 | | | 0.863 |
| | | | | | | 0.0003 | | | 0.850 |
| | | | | | | 0.0005 | | | 0.778 |
| | | | | | | | 16 | | 0.789 |
| | | | | | | | 32 | | 0.842 |
| | | | | | | | 128 | | 0.869 |
| | | | | | | | | BCE | 0.860 |

**Supplementary Figure 1:** Histograms of laboratory test results.

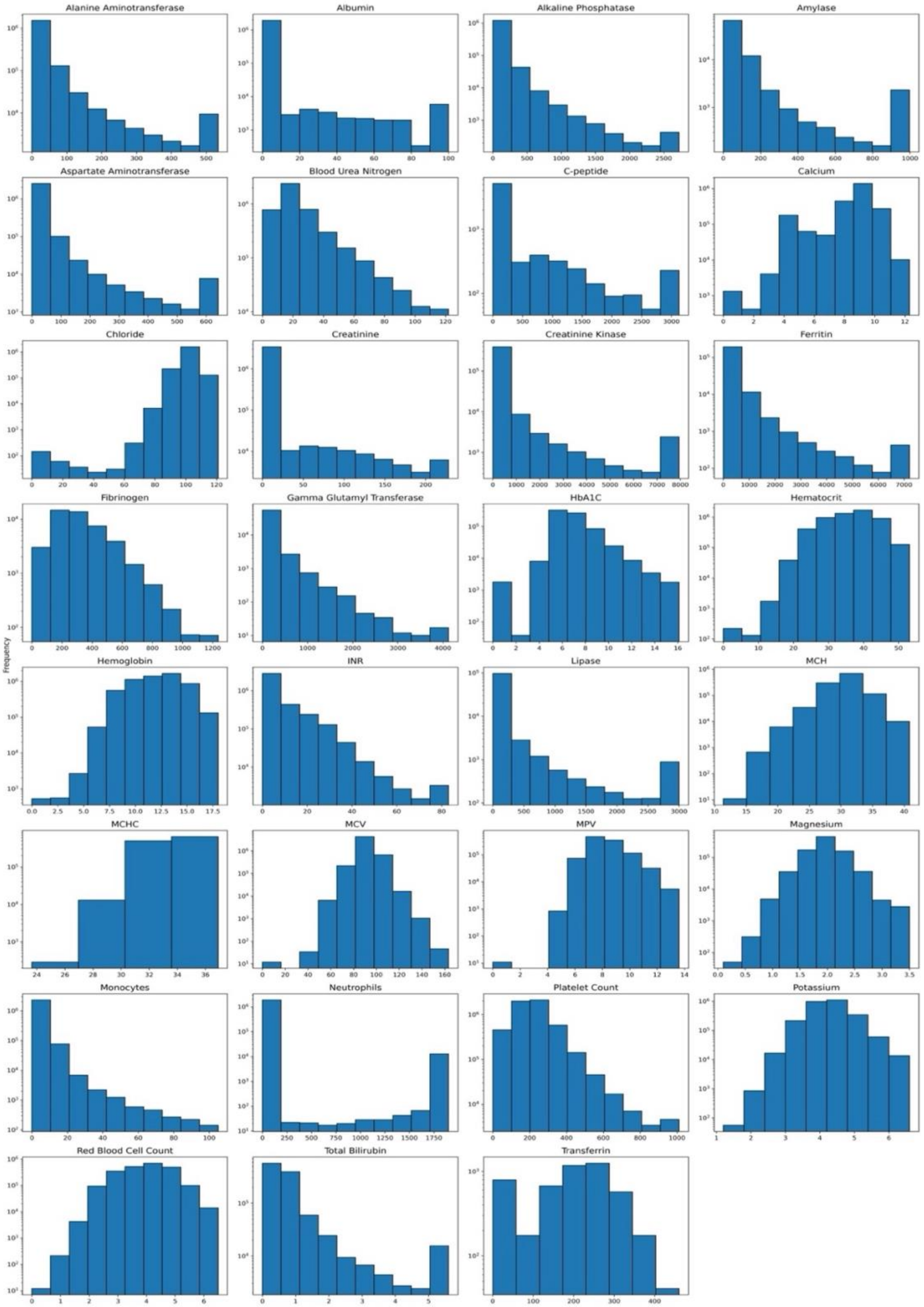

*International normalized ratio, INR; mean corpuscular hemoglobin, MCH; mean corpuscular hemoglobin concentration, MCHC; mean corpuscular volume, MCV; mean platelet volume, MPV.*

**Supplementary Figure 2:** Cohort selection flowcharts for case and control groups.

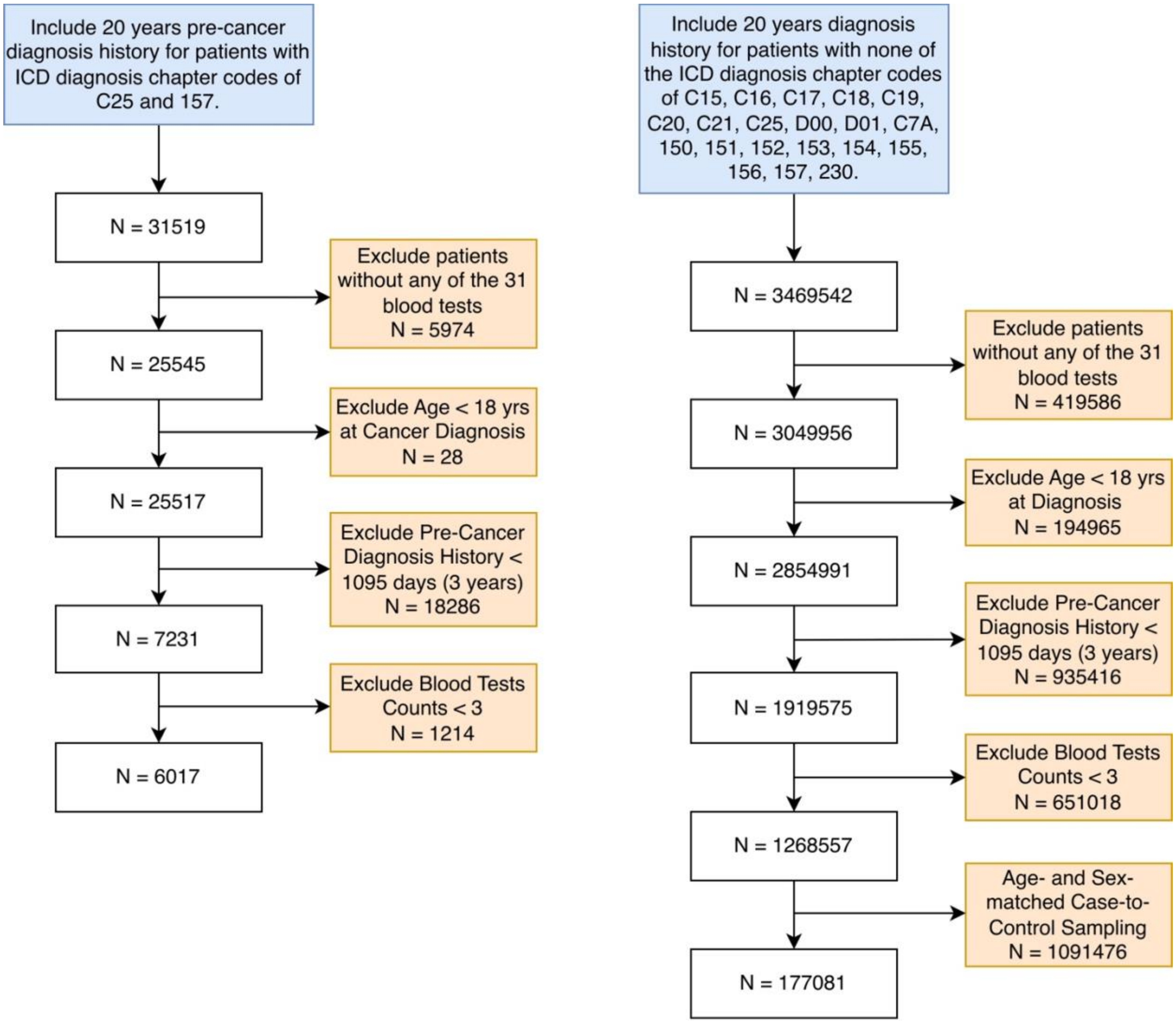

**Supplementary Figure 3**: Receiver operating characteristic curve for model including blood-based measure trajectories only (olive), medical trajectories only (purple), and the combined mode (black).

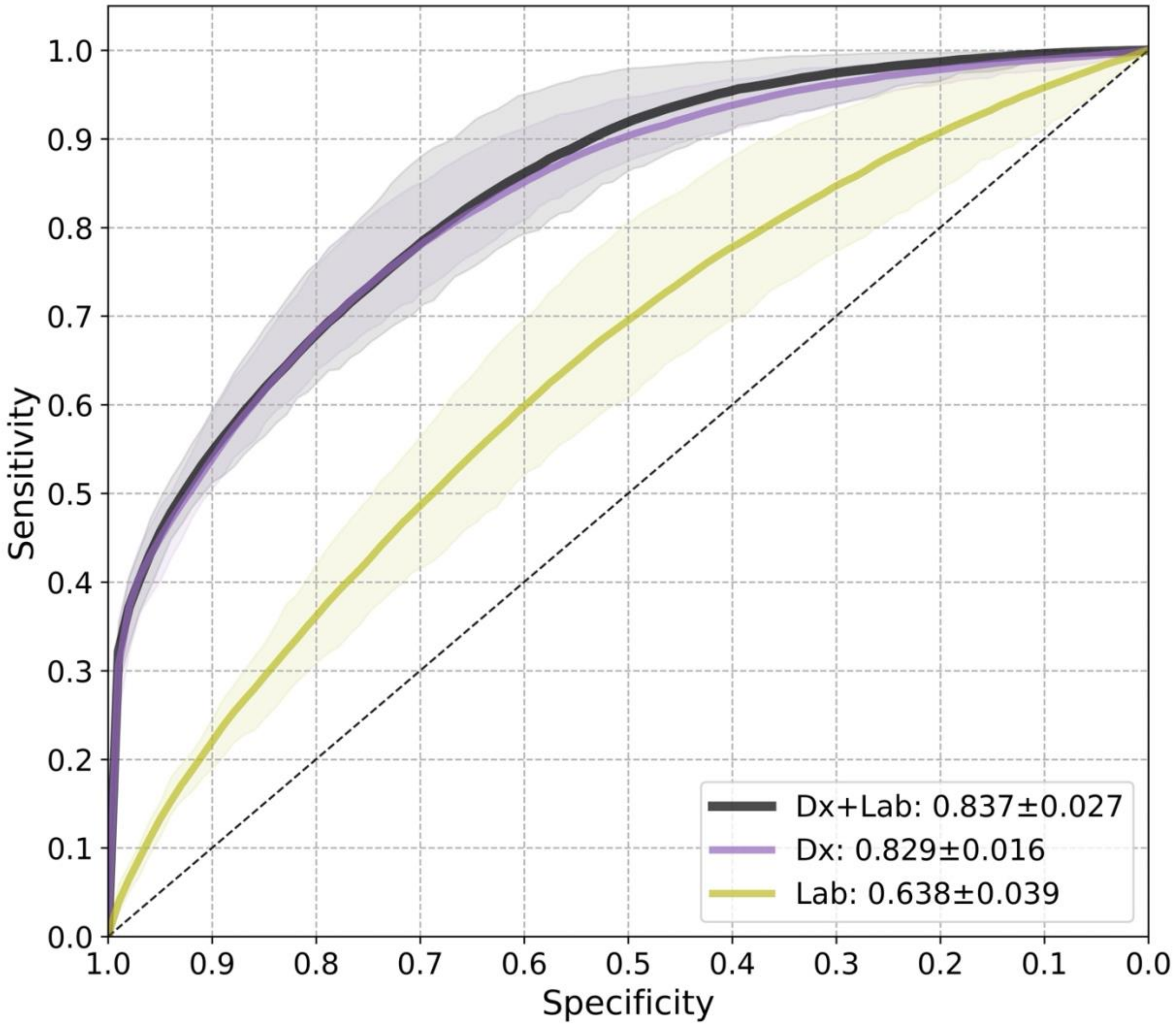

**Supplementary Figure 4**: Decision curve analysis of the trained PREMOD model on the held-out MCHS dataset with (A) 12-month, (B) 24-month, and (C) 36-month prediction lead times.

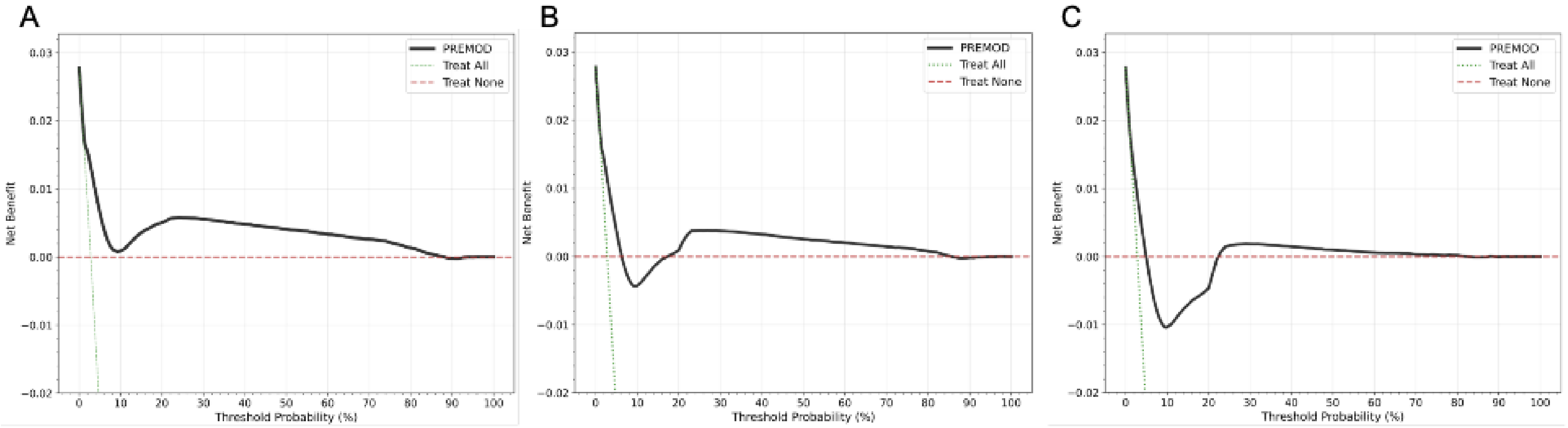